\documentclass[journal]{IEEEtran} 
\usepackage{amsmath,amsfonts}
\usepackage{array}
\usepackage{textcomp}
\usepackage{verbatim}
\usepackage{graphicx}
\usepackage{xcolor}
\usepackage{booktabs}
\usepackage{pgfplots}
\pgfplotsset{compat=1.18}
\usepackage{xcolor}
\usepackage[bookmarks=true]{hyperref}

\usepackage{hhline}
\usepackage{amsmath}  %
\usepackage[Symbol]{upgreek}%
\usepackage{bm}%
\usepackage{graphicx}
\usepackage{caption}
\usepackage{subfig}
\usepackage{epsfig}
\usepackage{mathrsfs}
\usepackage{amsfonts,amssymb}
\usepackage{algpseudocode}
\usepackage{cite}
\usepackage{supertabular}
\usepackage{longtable}
\usepackage{hhline}
\usepackage{multirow} 
\usepackage[normalem]{ulem}
\usepackage{dblfloatfix}    
\usepackage[bottom]{footmisc} 
\usepackage{tikz}
\usepackage[commandnameprefix=always]{changes}
\usepackage[T1]{fontenc}
\usepackage{aecompl}
\usepackage{mathtools}
\usepackage{booktabs}
\usepackage{tabularx}
\usepackage{threeparttable}
\usepackage{makecell}
\usepackage{rotating}   
\usepackage{cuted}  

\makeatletter

\newcommand{\Rmnum}[1]{\expandafter\@slowromancap\romannumeral #1@}

\makeatother

\usepackage[utf8]{inputenc}
\usepackage[english]{babel}
\usepackage{amsthm}

\hypersetup{hidelinks} 
\let\vec\boldvec
\let\mat\boldvec
\def\BibTeX{{\rm B\kern-.05em{\sc i\kern-.025em b}\kern-.08em
    T\kern-.1667em\lower.7ex\hbox{E}\kern-.125emX}}

\begin{document}

\title{Strategy-Supervised Autonomous Laparoscopic Camera Control via Event-Driven Graph Mining}

\author{Keyu~Zhou$^{\dagger}$,
        Peisen~Xu$^{\dagger}$,
        Yahao~Wu,
        Jiming~Chen$^{*}$,
        Gaofeng~Li$^{*}$,
        and~Shunlei~Li$^{*}$%
\thanks{$^{\dagger}$Keyu Zhou and Peisen Xu contributed equally to this work.}%
\thanks{Keyu Zhou, Yahao Wu, Jiming Chen, and Shunlei Li are with Hangzhou Dianzi University, Hangzhou 310018, China (e-mail: jmchen@hdu.edu.cn; shunlei.li@outlook.com).}%
\thanks{Peisen Xu and Gaofeng Li are with Zhejiang University, Hangzhou 310058, China (e-mail: gaofeng.li@zju.edu.cn).}%
\thanks{$^{*}$Corresponding authors: Jiming Chen, Gaofeng Li, and Shunlei Li.}
}


\maketitle

\begin{abstract}
Autonomous laparoscopic camera control must maintain a stable and safe surgical view under rapid tool--tissue interactions while remaining interpretable to surgeons. We present a strategy-grounded framework that couples high-level vision-language inference with low-level closed-loop control. Offline, raw surgical videos are parsed into camera-relevant temporal events (e.g., interaction, working-distance deviation, and view-quality degradation) and structured as attributed event graphs. Mining these graphs yields a compact set of reusable camera-handling strategy primitives, which provide structured supervision for learning. Online, a fine-tuned Vision--Language Model (VLM) processes the live laparoscopic view to predict the dominant strategy and discrete image-based motion commands, executed by an IBVS--RCM controller under strict safety constraints; optional speech input enables intuitive human-in-the-loop conditioning. On a surgeon-annotated dataset, event parsing achieves reliable temporal localization (F1-score 0.86), and the mined strategies show strong semantic alignment with expert interpretation (cluster purity 0.81). Extensive \textit{ex vivo} experiments on silicone phantoms and porcine tissues demonstrate that the proposed system outperforms junior surgeons in standardized camera-handling evaluations, reducing field-of-view centering error by 35.26\% and image shaking by 62.33\%, while preserving smooth motion and stable working-distance regulation.
\end{abstract}


\section{Introduction}
\IEEEPARstart{M}{inimally} Invasive Surgery (MIS) has fundamentally transformed the landscape of modern medicine. By minimizing tissue trauma and accelerating postoperative recovery, it has established itself as the gold standard for thoracic, abdominal, and pelvic interventions \cite{dupont2021decade, silvera2024robotics}. Within this paradigm, robotic laparoscopy represents a critical category where  endoscopes and instruments are inserted through small incisions (trocars). In these procedures, the laparoscopic camera functions as the surgeon's sole visual interface, making stable, centered, and optimal Field-of-View (FoV) management a non-negotiable prerequisite for safety \cite{maier2017surgical}. Traditionally, the laparoscope is manipulated by a human assistant. However, this human-in-the-loop approach is inherently constrained: physiological fatigue, hand tremors, and inevitable miscommunication can disrupt the surgical workflow, leading to unstable views and increased cognitive load for the operating surgeon \cite{pandya2014review, cao2016pupil}. These limitations are exacerbated in long-duration procedures, catalyzing the urgent demand for robotic laparoscopic camera holders designed to augment surgical capabilities with intelligent autonomy \cite{sandoval2019autonomous, attanasio2021autonomy}.

To automate camera guidance, foundational research predominantly relied on classical Visual Servoing (VS) techniques. These systems generally employ Image-Based Visual Servoing (IBVS) control laws to maintain surgical instruments within the FoV by minimizing feature errors in the image plane \cite{chaumette2006visual, krupa2003autonomous}. To address the mechanical constraints of surgical ports, subsequent works integrated Remote Center of Motion (RCM) constraints into the control loop to ensure safe manipulator movements \cite{zhou2018new, aghakhani2013task}. Furthermore, to handle the dynamic uncertainties inherent in the abdominal environment—such as physiological motion (e.g., respiration) and unmodeled disturbances during instrument-tissue interaction—researchers have adopted adaptive and model-free control schemes \cite{wang2023cerebellum, ott2011robotic, zhang2023visual}. Despite their geometric precision, these VS methods remain inherently reactive. They treat the laparoscope merely as a "follower" of the tool tip without comprehending the surgical context. This often results in jittery, unsafe motions when tools move rapidly or inadvertently, and fails to handle scenarios where the visual target is temporarily occluded or exits the view \cite{agustinos2014visual, peng2022endoscope}.

The advent of Deep Learning (DL) has significantly elevated the perception capabilities of robotic systems, attempting to address the robustness issues of traditional feature trackers. Convolutional Neural Networks (CNNs) have enabled robust instrument segmentation and detection even under challenging illumination and occlusion scenarios \cite{allan20192017, shvets2018automatic}. Building on these perception backbones, recent state-of-the-art frameworks have focused on handling the deformability of the surgical environment. For example, EndoTracker \cite{seenivasan2025endotracker} introduced a robust point tracking mechanism that maintains consistency across deformable tissue changes, while other works have explored multi-source fusion to resolve depth ambiguity in monocular laparoscopic views \cite{luo2024multisource}. Moreover, specific functionalities like automatic zoom control based on tool geometry have been developed; Zhang et al. \cite{zhang2025model} recently demonstrated a model predictive control framework that adjusts the zoom factor to maintain consistent tool tracking. However, improved perception does not equate to intelligent control. Most existing systems still map visual detections directly to kinematic velocities in an end-to-end manner. They lack a temporal and semantic understanding of the procedure—knowing "where" the tool is, but not what the surgeon is doing—leading to a system that cannot anticipate future needs or distinguish between critical maneuvers and incidental movements \cite{banks2025autocam}.

To bridge the gap between reactive tracking and cognitive autonomy, the research frontier has shifted towards integrating surgical workflow recognition with high-level planning. By identifying surgical phases or action triplets, robots can theoretically adapt their control policies to specific procedural requirements \cite{padoy2019machine, czempiel2020tecno}. Reinforcement Learning (RL) and Imitation Learning (IL) have been widely applied to learn camera policies directly from expert demonstrations \cite{eslamian2020development, osa2018algorithmic}. A notable advancement is AutoCam \cite{banks2025autocam}, which proposes a hierarchical path planning framework that decouples strategic viewpoint positioning from local obstacle avoidance, allowing for more deliberate camera placements. Similarly, diffusion Policies \cite{chi2025diffusion} have emerged as a powerful tool for modeling complex, multi-modal action distributions, showing promise in benchmarks like SutureBot \cite{suturebot} for autonomous tissue manipulation. Nevertheless, translating discrete phase recognition or learned distributions into continuous, smooth control remains a formidable challenge. Most learning-based black box models struggle with generalization across different patients, unexpected anatomical variations, or unseen instrument types, often failing to maintain the strategic consistency required for safety-critical execution \cite{paul2024voice, valderrama2022towards}.

The paradigm shift towards Large Language Models (LLMs) and Vision-Language Models (VLMs) offers a transformative opportunity for Embodied AI. Generalist foundation models like RT-2 \cite{zitkovich2023rt}, PaLM-E \cite{driess2023palme}, and the recent Gemini Robotics framework \cite{gemini} have demonstrated an unprecedented ability to ground high-level language instructions into low-level robotic actions, enabling zero-shot generalization to new tasks \cite{openvla}. In the medical domain, this has inspired systems like EndoVLA \cite{ng2025endovla}, which utilizes dual-phase vision-language-action models for precise tracking, and RoboNurse-VLA \cite{robonurse}, which acts as a robotic scrub nurse. Furthermore, SurgeryLLM \cite{surgeryllm} and other retrieval-augmented frameworks \cite{srt} attempt to provide decision support by leveraging vast textual medical knowledge. 
Despite this promise, the direct application of general-purpose VLMs to laparoscopic control is hindered by a significant domain gap. A generic model does not inherently understand the subtle, strategy-laden "grammar" of surgical camera handling—such as maintaining the horizon, anticipating tool workspace, or prioritizing safety zones \cite{ahn2022can, zhang2025llm}. Moreover, the probabilistic nature of generative models poses safety risks such as hallucination, necessitating rigorous grounding mechanisms \cite{shieldagent}. Consequently, there is a critical need for systems that can explicitly discover and utilize these latent surgical strategies—essentially "mining" the tacit knowledge of expert camera operators—to supervise and constrain the generative capabilities of large models.

\begin{figure*}[htbp]
    \centering
    \includegraphics[width=0.95\textwidth]{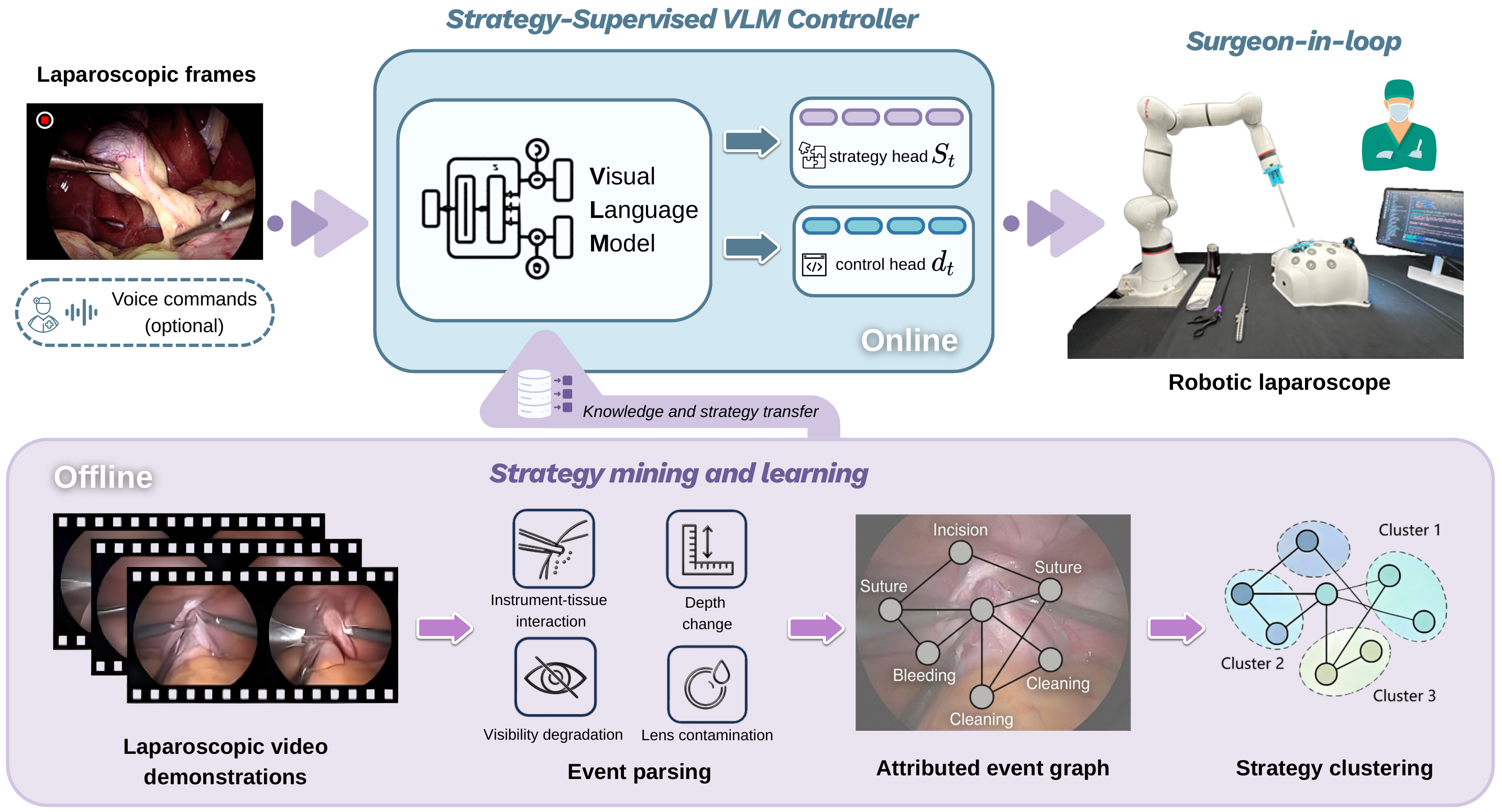}
    \caption{\textbf{Overview of the proposed strategy-supervised autonomous laparoscopic camera control framework:} The bottom block illustrates the offline pipeline, where laparoscopic video demonstrations are parsed into camera-relevant events, organized as attributed event graphs, and clustered to discover  dominant camera-handling strategy primitives. These clusters generate strategy labels and corresponding direction supervision signals. The top row shows the online control pipeline: a laparoscopic frame (with optional speech input) is processed by a vision-language backbone with two prediction heads that output a dominant strategy label and discrete 6-DoF motion directions. The predicted directions are executed by an IBVS–RCM controller to maintain stable, safe, and context-aware camera positioning, with surgeon-in-the-loop safety override.}

    \label{fig:frame}
\end{figure*}

To address these limitations, we propose a strategy-grounded autonomous laparoscopic camera control framework that couples interpretable strategy discovery with safety-constrained real-time execution, as illustrated in Fig. \ref{fig:frame}. Rather than attempting to directly regress continuous camera velocities from raw pixels---a process often fraught with instability and lack of interpretability---we adopt a hierarchical approach. \textbf{Offline}, we transform raw surgical videos into an event-level representation that captures \emph{what} surgical action is occurring and \emph{why} a specific viewpoint is preferred. By modeling these events via an attributed graph formulation, we mine recurring camera-handling patterns, yielding a compact set of reusable \textbf{strategy primitives} that serve as high-level supervision. \textbf{Online}, a multi-modal policy leverages the live laparoscopic view, the inferred strategy context, and optional surgeon voice commands to output discrete motion intentions. These intentions are subsequently executed by a safety layer that enforces kinematic, workspace, and Remote Center of Motion (RCM) constraints. This design explicitly decouples (i) perception and event abstraction, (ii) strategy reasoning, and (iii) low-level control, ensuring robust behavior even under rapid tool motions, transient occlusions, and inter-procedure variability.

The main contributions of this article are summarized as follows:
\begin{itemize}
    \item \textbf{Strategy-grounded camera control pipeline:} We introduce a hierarchical autonomous control framework that extracts \emph{explicit} camera-handling strategies from expert demonstrations to guide closed-loop execution, effectively bridging the gap between reactive visual servoing and black-box end-to-end imitation.
    \item \textbf{Event abstraction and attributed strategy mining:} We propose an event-centric representation of surgical workflow and an attributed graph mining approach that discovers reusable strategy primitives by jointly leveraging temporal, visual, kinematic, and semantic cues.
    \item \textbf{Multi-modal, strategy-conditioned policy with safety constraints:} We develop a VLM-based policy that fuses endoscopic observations with strategy context and optional voice commands. This policy is integrated with a rigorous safety layer to guarantee smooth, RCM-compliant motion suitable for clinical environments.
    \item \textbf{Real-time system validation:} We implement the full pipeline on a robotic laparoscope holder and validate it via comprehensive \textit{ex vivo} studies (including porcine tissue dissection and phantom suturing). Results demonstrate superior stability and visibility maintenance compared to manual operation and baseline methods.
\end{itemize}

Overall, this work demonstrates that incorporating mined behavioral strategies as supervision creates a structured and transparent control paradigm. By coupling strategy-supervised prediction with an IBVS--RCM controller, our system achieves stable closed-loop performance across diverse events---ranging from routine depth adjustments to handling visibility degradation (e.g., smoke or lens contamination). Furthermore, the framework's support for multi-modal intervention provides a practical foundation for interpretable, surgeon-collaborative robotic assistance.

\section{Related Works}

\subsection{Event-Based Modeling in Surgical Video}
\label{sec:rw_event_video}

A large body of work structures surgical videos by recognizing procedure-level
workflow or phase transitions, providing a coarse yet clinically meaningful
temporal abstraction. Early deep learning approaches such as EndoNet jointly
modeled phase recognition and tool presence, demonstrating the benefit of
multi-task visual cues for laparoscopic workflow understanding~\cite{twinanda2016endonet}.
To better handle long untrimmed videos, subsequent methods emphasized explicit
temporal modeling and online refinement. Representative examples include
multi-stage temporal convolutional networks (e.g., TeCNO) that progressively
refine predictions over time~\cite{czempiel2020tecno}, and Transformer-based
formulations that aggregate spatio-temporal embeddings for accurate phase
recognition (e.g., Trans-SVNet)~\cite{gao2021trans}. More recent work further
improves long-horizon temporal reasoning via auto-regressive dependencies
(e.g., ARST)~\cite{zou2023arst} and efficient key-information pooling for
online recognition (e.g., SKiT)~\cite{liu2023skit}, while long-video
Transformers such as LoViT explicitly target the computational and ambiguity
challenges of long laparoscopic recordings~\cite{liu2025lovit}.

Beyond phase-level abstraction, fine-grained event/activity modeling has gained
increasing attention, aiming to capture \emph{what} the surgeon is doing through
tool-centric action semantics and interaction patterns. In particular, action
triplet modeling (tool--verb--target) has been promoted through community
benchmarks such as CholecTriplet, extending recognition to detection to provide
more localized context in endoscopic scenes~\cite{nwoye2023cholectriplet2022}.
Related community challenges on workflow recognition in endoscopic procedures
further reflect the growing interest in structured temporal understanding across
diverse surgical domains~\cite{das2024pitvis}.

While these studies provide strong tools for semantic recognition of phases,
activities, and interactions, they primarily output categorical interpretations
of surgical progress or actions. In contrast, our goal is to extract
\emph{camera-relevant} event intervals and encode them with behaviorally
meaningful descriptors of both situational cues and interval-level camera
responses, enabling downstream discovery of reusable camera-handling strategy
primitives rather than predefined semantic labels.

\subsection{Graph-Based Strategy Discovery and Clustering}
\label{sec:rw_wsbgc}

Graph-based clustering provides a natural tool for discovering latent structure
from relational data, especially when meaningful patterns are induced jointly by
connectivity and node attributes. In attributed settings, coupling topology with
node features has been shown to improve community discovery and robustness to
heterogeneity~\cite{yang2013community}, and has motivated a broad class of methods
that encode structure and features in a unified representation space~\cite{kipf2016semi}.
When multiple, complementary relations are available (e.g., temporal adjacency
and attribute affinity), multi-view fusion further improves clustering stability
by integrating multiple graphs rather than relying on a single noisy relation
source~\cite{nie2017self}.

A representative formulation that combines these principles is the Weighted
Symmetric Boosted Graph Clustering (WSBGC) framework~\cite{li2025weighted}, which
constructs boosted and refined affinity graphs before performing weighted symmetric
NMF-based clustering. WSBGC first integrates temporal and attribute-based relations
into an auxiliary boosted graph:
\begin{equation}
    \mathbf{G}_{A} =
    \mu \mathbf{S} + (1 - \mu)\mathbf{A},
    \label{eq:GA}
\end{equation}
where $\mathbf{A}$ denotes temporal adjacency, $\mathbf{S}$ denotes attribute similarity,
and $\mu \in [0,1]$ balances their contributions.

To capture latent functional relations beyond raw proximity, WSBGC adopts a
self-expressiveness mechanism from subspace clustering~\cite{elhamifar2013sparse}:
\begin{equation}
    \min_{\mathbf{C} \ge 0}
    \;
    \lVert \mathbf{X} - \mathbf{X}\mathbf{C} \rVert_F^2
    + \alpha \lVert \mathbf{C} \rVert_1,
    \label{eq:selfexpress}
\end{equation}
where $\mathbf{X}$ is the attribute matrix and $\alpha$ controls sparsity.
The coefficients are symmetrized to obtain:
\begin{equation}
    \mathbf{S}_{c} =
    \frac{1}{2}(\mathbf{C} + \mathbf{C}^{\top}).
    \label{eq:Sc}
\end{equation}
A refined graph is then formed by fusing learned relations with the original
attribute affinity:
\begin{equation}
    \mathbf{G}_{R} =
    \lambda \mathbf{S}_{c} + (1 - \lambda)\mathbf{S},
    \label{eq:GR}
\end{equation}
where $\lambda \in [0,1]$ controls the contribution of learned relational information.

Finally, clustering is achieved via weighted symmetric NMF, which is closely related
to spectral clustering in graph settings~\cite{ding2005equivalence}:
\begin{equation}
    \min_{\mathbf{U} \ge 0}
    \;
    \frac{1}{2}
    \lVert
        \mathbf{W} \odot (\mathbf{G}_{R} - \mathbf{U}\mathbf{U}^{\top})
    \rVert_F^2,
    \label{eq:WSNMF}
\end{equation}
where $\mathbf{U}$ encodes soft cluster memberships and $\mathbf{W}$ assigns confidence
weights to graph entries.

Such graph-boosted formulations are effective when latent modes are driven jointly
by workflow structure and attribute-level consistency, aligning with our goal of
mining reusable camera-handling strategy primitives from surgical event graphs.
However, existing graph-based clustering approaches are typically designed for
static pattern discovery and do not directly address the requirements of
closed-loop medical robotic control. In laparoscopic camera guidance, the mined
modes must be temporally stable, interpretable to surgeons, and compatible with
RCM-constrained visual servoing under strict safety constraints. Moreover,
surgical scenes exhibit highly dynamic tool--tissue interactions, frequent
visibility degradation, and ambiguous event boundaries, which challenge
conventional clustering methods that assume clean and stationary embeddings.
Bridging graph-level structure mining with real-time, safety-critical
endoscopic control therefore requires a task-aware formulation that links
event semantics to executable motion supervision, motivating the proposed
strategy-grounded framework.

\section{Methods}

This section presents a strategy-grounded framework for autonomous
laparoscopic camera control.
From surgical video demonstrations, we extract camera-relevant events and
encode them into an attributed event graph capturing temporal context,
visual conditions, and interval-level camera responses.
Clustering event instances in this joint state--action space yields recurrent
camera-handling primitives and their discrete direction prototypes, which
supervise a real-time visual model that predicts camera motion direction from
intraoperative images.
Predicted directions are executed by a classical IBVS--RCM controller, which determines motion magnitude while enforcing
remote-center-of-motion constraints and surgical safety limits.

The pipeline is organized as follows:
\begin{itemize}
\item \textbf{A--B: Offline Event Parsing and Attributed Graph Construction.}
Videos are decomposed into temporally contiguous events from three categories:
interaction-driven, depth-change, and view-quality constraints.
Each event is augmented with multi-modal descriptors (tool kinematics,
deformation, depth, visibility) and interval-level camera-response statistics,
forming an attributed graph $(A,X,S)$ with temporal adjacency and semantic
affinity.

\item \textbf{C--D: Graph-Boosted Clustering and Strategy Discovery.}
Following WSBGC, we cluster events in the graph-informed space to uncover
latent state--action modes. Each cluster defines a reusable strategy primitive
and a discrete 6D direction prototype.

\item \textbf{E: Strategy-Supervised Direction Prediction.}
Direction labels are propagated to all frames within their event intervals.
A visual model predicts $\mathbf{d}_t \in \{-1,0,+1\}^6$ from images, with
auxiliary strategy supervision. At runtime, $\mathbf{d}_t$ is forwarded to the
IBVS--RCM controller for safe closed-loop execution.
\end{itemize}

Together, these components provide a reproducible pathway from surgical videos
to a deployable, strategy-consistent camera-control system operating in real
time.

\subsection{Offline Event Parsing}

\label{sec:offline_event_parsing}

\begin{figure*}[htbp]
    \centering
    \includegraphics[width=0.95\textwidth]{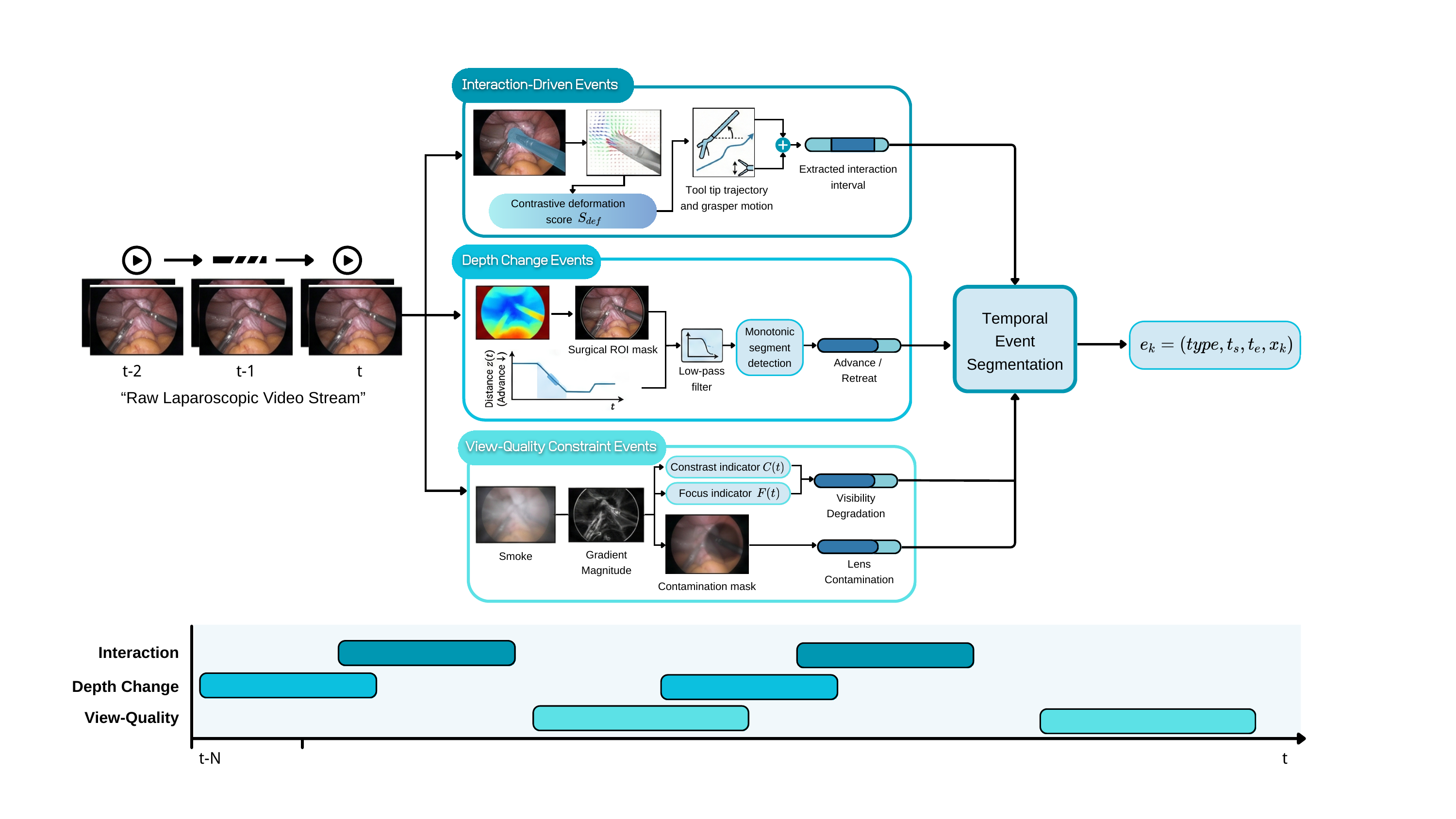}
    \caption{\textbf{Event parsing from raw laparoscopic video:} Three branches detect interaction-driven events, depth-change events, and view-quality constraint events (visibility degradation and lens contamination), whose candidate intervals are fused by temporal event segmentation to form structured events $e_k=(\text{type}, t_s, t_e, x_k)$. The bottom timeline illustrates the resulting aligned event segments over time.}

    \label{fig:event}
\end{figure*}

The first stage of the proposed system converts raw laparoscopic videos into a
structured, camera-relevant representation that supports subsequent strategy
mining and multi-modal instruction learning as shown in Fig. \ref{fig:event}. 
We localize temporally contiguous
event instances by analyzing cues that directly correlate with expert camera
handling, including primary-tool motion with induced local tissue deformation,
global depth-motion cues reflecting camera advance/retreat, and view-quality
indicators capturing smoke/fog and lens contamination. Each event is time-indexed
and augmented with multi-modal attributes, including behavioral descriptors of
the camera response observed over the same interval (e.g., re-centering magnitude
and depth adjustment), enabling strategy mining in a behaviorally meaningful space.

Laparoscopic scenes exhibit substantial variability in anatomy, illumination,
and tool--tissue interaction patterns. While one could enumerate many
fine-grained micro-events, doing so would yield a fragmented and procedure-specific
representation with limited robustness and poor utility for discovering reusable
camera-handling strategies.

To balance expressiveness and generalizability, we adopt a compact event taxonomy
with three camera-relevant classes: \textbf{interaction-driven events} capturing
effective instrument--tissue engagement, \textbf{depth change events} capturing
camera motion along the optical axis, and \textbf{view-quality constraint events}
capturing transient visibility degradation and lens contamination. This taxonomy
is intentionally minimal and separates distinct causal mechanisms that require
qualitatively different camera responses. Finer-grained variations are encoded
through continuous event attributes and camera-response descriptors rather than
additional discrete labels. Finally, adjustments that depend on explicit surgeon
intent and are not reliably inferable from vision alone are incorporated via an
optional speech-modifier channel.

\subsubsection{Interaction-Driven Events}

Interaction-driven events are identified by correlating primary tool kinematics with the resulting spatio-temporal tissue response. This coupling allows us to distinguish purposeful manipulation from idle movements or camera-induced motion.

The primary surgical instrument is segmented and tracked to extract the tool-tip trajectory $p_t$ and grasper aperture $\theta_t$. To quantify the mechanical effect on tissue, we compute a dense optical flow field $\mathbf{u}(x,y,t)$ between consecutive frames. To isolate local manipulation from global disturbances (e.g., respiration or camera ego-motion), we define a \textit{contrastive deformation score} $S_{\mathrm{def}}(t)$:
\begin{equation}
    S_{\mathrm{def}}(t) = \frac{\mathrm{median}_{(x,y)\in R_{\mathrm{near}}(t)} \|\mathbf{u}(x,y,t)\|_2}{\mathrm{median}_{(x,y)\in R_{\mathrm{ref}}(t)} \|\mathbf{u}(x,y,t)\|_2 + \epsilon},
\end{equation}
where $R_{\mathrm{near}}(t)$ is a tool-centric adaptive neighborhood (defined as a disk centered at $p_t$ with radius $r_{\mathrm{act}}$), and $R_{\mathrm{ref}}(t)$ represents distal tissue regions representing the global motion floor. 
The tool-centric near-region radius was set to $r_{\mathrm{act}}=60$ pixels for 30 pixels after resizing the shorter side to 512, which empirically balances capturing local tool–tissue deformation while suppressing global motion effects.
To ensure data integrity, frames identified with severe view-quality constraints are treated as occlusion gaps and handled via temporal interpolation or event truncation.

To suppress stochastic noise and artifacts from specular reflections, the signal $S_{\mathrm{def}}(t)$ is smoothed using a Savitzky-Golay filter. An interaction event is declared over a maximal interval $[t_s, t_e]$ when the following conditions are simultaneously satisfied:

    (1) Tissue Response: $S_{\mathrm{def}}(t) > \tau_{\mathrm{def}}$ for $t \in [t_s, t_e]$, ensuring significant local deformation relative to the background.
    
    (2) Kinematic Intent: The instrument exhibits active motion or manipulation, expressed as:
    \begin{equation}
        \|\dot{p}_t\| > \tau_p \quad \lor \quad |\dot{\theta}_t| > \tau_\theta,
    \end{equation}
    where $\dot{p}_t$ and $\dot{\theta}_t$ denote the temporal derivatives of position and grasper angle.
    
    (3) Sustained Engagement: The interval must satisfy $(t_e - t_s) \ge T_{\min}$ and a cumulative deformation energy constraint $\sum_{t=t_s}^{t_e} S_{\mathrm{def}}(t) > \Delta_{\mathrm{def}}$, filtering out transient contacts or jitter.

Hysteresis-based bridging is applied to $S_{\mathrm{def}}(t)$ to maintain event continuity during brief periods of instrument stasis within a single manipulation task.

\subsubsection{Depth Change Events}

Depth change events characterize the laparoscope's translational motion along the optical axis, altering the camera-to-scene distance. These maneuvers are fundamental to surgical workflow, marking transitions between global exploration and localized fine-grained tasks.

For each frame, we estimate a dense depth map $D_t(u,v)$ using Surgical-DINO \cite{cui2024surgical}, a foundation model optimized for the unique photometric challenges of endoscopic environments, such as specular reflections and non-Lambertian tissue surfaces. To derive a stable representation of the camera's proximity to the workspace, we compute a scalar depth signal $z(t)$ as the median value within the surgical region-of-interest $R_{\mathrm{surg}}(t)$:
\begin{equation}
    z(t) = \mathrm{median}_{(u,v)\in R_{\mathrm{surg}}(t)} D_t(u,v).
\end{equation}
The median operator provides inherent robustness against localized depth estimation failures and instrument occlusions. To suppress high-frequency jitter (e.g., physiological tremor) and sensor noise, $z(t)$ is processed through a low-pass temporal filter with a cutoff frequency $f_c$ of 2--3\,Hz. The resulting filtered signal $\tilde{z}(t)$ captures the persistent trend of the camera’s ego-motion.

A depth change event is triggered when $\tilde{z}(t)$ exhibits a sustained monotonic trend. We compute the temporal derivative $\dot{\tilde{z}}(t)$ and declare an event over a maximal interval $[t_s, t_e]$ that satisfies:
\begin{equation}
    |\dot{\tilde{z}}(t)| > \tau_{z} \quad \text{for} \quad t\in[t_s, t_e], \qquad (t_e-t_s) \ge T_{\min}.
\end{equation}
To distinguish purposeful camera navigation from residual drift, we further require the cumulative displacement to exceed a minimum magnitude: $|\tilde{z}(t_e) - \tilde{z}(t_s)| > \Delta_{\min}$. 
The minimum cumulative depth change threshold was defined as $\Delta_{\min}=0.08\,\mathrm{median}(z)$, providing a scale-adaptive criterion that remains robust across different anatomical scenes.

Events are classified as \textit{Advance} ($\dot{\tilde{z}} < 0$) or \textit{Retreat} ($\dot{\tilde{z}} > 0$). By integrating these events with tool--tissue interaction data, the system can subsequently distinguish between active zooming for precise dissection and exploratory withdrawal for situational awareness.

\subsubsection{View-Quality Constraint Events}

View-quality constraint events capture scenarios where visual perception is degraded, necessitating camera adjustment or cleaning. We distinguish between two operationally distinct cases: transient visibility degradation (e.g., surgical smoke) and persistent lens contamination (e.g., blood or fat splatters).

\paragraph{Transient visibility degradation (smoke/fog).}

Surgical smoke introduces global contrast attenuation and spatial blurring. To ensure robustness against varying illumination and automatic gain control (AGC), we employ normalized statistical measures. We monitor the focus measure $F(t)$ and a scale-invariant contrast measure $C(t)$ over the surgical region $R_{\mathrm{surg}}(t)$.

The focus measure is defined by the median spatial gradient magnitude to suppress localized noise:
\begin{equation}
    F(t) = \mathrm{median}_{(u,v)\in R_{\mathrm{surg}}(t)} \|\nabla I_t(u,v)\|_2,
\end{equation}
where $I_t$ is the image intensity. To decouple contrast from absolute brightness, we define $C(t)$ as the coefficient of variation:
\begin{equation}
    C(t) = \frac{\sigma(I_t)}{\mu(I_t)} \quad \forall (u,v) \in R_{\mathrm{surg}}(t),
\end{equation}
where $\sigma(\cdot)$ and $\mu(\cdot)$ denote the standard deviation and mean intensity, respectively. A degradation event is triggered over an interval $[t_s, t_e]$ if $F(t) < \tau_F$ and $C(t) < \tau_C$ persist for a duration $\ge T_{\min}$, with hysteresis applied to prevent oscillation near threshold boundaries. 
The minimum event duration was fixed to $T_{\min}=0.6$\,s, which effectively filters transient jitter and brief occlusions while preserving meaningful surgical interactions.

\paragraph{Lens contamination.}

Lens contamination produces localized occlusions that remain spatially fixed relative to the camera sensor. Let $B_t(u,v) \in \{0,1\}$ be a binary low-visibility map obtained by thresholding the local gradient energy. To distinguish static lens artifacts from dynamic scene content, we enforce a \textit{spatio-temporal consistency constraint}.

We define a temporal accumulation map $\bar{B}_t$ over a sliding window of 
size $K_w$:
\begin{equation}
    \bar{B}_t(u,v) =
    \frac{1}{K_w}
    \sum_{k=0}^{K_w-1}
    B_{t-k}(u,v).
\end{equation}
The contamination persistence window was set to $K_w=0.8$\,s (approximately 24 frames at 30\,fps), ensuring stable detection of sensor-fixed artifacts while avoiding false positives from transient low-texture regions.
The contamination score $S_{\mathrm{cont}}(t)$ is then computed as the fraction of pixels exhibiting high temporal persistence:
\begin{equation}
    S_{\mathrm{cont}}(t) = \frac{1}{|R_{\mathrm{surg}}(t)|} \sum_{(u,v)\in R_{\mathrm{surg}}(t)} \mathbb{I}(\bar{B}_t(u,v) > \tau_{\mathrm{persist}}),
\end{equation}
where $\mathbb{I}(\cdot)$ is the indicator function and $\tau_{\mathrm{persist}}$ is the persistence threshold (typically set to $0.9$ to filter out moving surgical tools). 

An event is declared when $S_{\mathrm{cont}}(t) > \tau_{\mathrm{cont}}$. 
The contamination ratio threshold was fixed to $\tau_{\mathrm{cont}}=0.12$, which reliably captures clinically relevant lens contamination without triggering on minor local contrast fluctuations.
Post-hoc validation is performed by detecting the abrupt restoration of $F(t)$ and $C(t)$ following a detected camera withdrawal, providing a secondary confirmation of successful lens cleaning.

\subsubsection{Event Representation and Normalization}
\label{sec:event_representation}

To facilitate attributed-graph construction and downstream strategy mining, all
detected instances are converted into a unified spatio-temporal representation.
Each event $e_k$ is defined as a tuple encoding its categorical, temporal, and
descriptive properties:
\begin{equation}
    e_k = \bigl( y_k, \ t_k^{s}, \ t_k^{e}, \ \mathbf{x}_k \bigr),
\end{equation}
where $y_k \in \mathcal{Y}$ is the discrete event label from our taxonomy,
$[t_k^{s}, t_k^{e}]$ denotes the temporal interval, and
$\mathbf{x}_k \in \mathbb{R}^d$ is a multi-modal attribute vector.

Importantly, $\mathbf{x}_k$ summarizes both the surgical situation (\emph{state})
and the observed laparoscope adjustment over the same interval (\emph{action}):
\begin{equation}
\mathbf{x}_k = \bigl[ \mathbf{x}^{\mathrm{state}}_k \ \| \ \mathbf{x}^{\mathrm{action}}_k \bigr].
\end{equation}
The state component $\mathbf{x}^{\mathrm{state}}_k$ encodes task-relevant cues
from interaction, depth, and view-quality signals, whereas the action component
$\mathbf{x}^{\mathrm{action}}_k$ encodes interval-level camera-response
descriptors in the image space. Concretely, we estimate frame-wise
camera-induced background motion between consecutive frames (excluding tool
regions and low-visibility pixels) and robustly fit a global 2D translation
model via RANSAC to obtain in-plane increments $(\delta u_t,\delta v_t)$.
Axial adjustment is derived independently from the filtered working-distance
trajectory $\tilde{z}(t)$, with $\delta z_t=\tilde{z}(t\!+\!1)-\tilde{z}(t)$.
These increments are aggregated over $[t_k^{s}, t_k^{e}]$ into signed
interval-level displacements $(\Delta u_k,\Delta v_k,\Delta z_k)$.
Full implementation details are provided in Appendix~A.

To ensure a fixed-length input for graph-based modeling, $\mathbf{x}_k$
concatenates the following descriptors, with non-applicable dimensions masked
per event type:
\begin{itemize}
    \item \textbf{Kinematic \& activity statistics:} temporal summaries (mean,
    peak, variance) of tool-tip velocity $\|\dot{p}_t\|$ and grasper actuation
    $|\dot{\theta}_t|$.
    \item \textbf{Biomechanical response:} interval statistics of the contrastive
    deformation signal $S_{\mathrm{def}}(t)$, capturing manipulation intensity.
    \item \textbf{Optical navigation descriptors:} for depth events, direction
    indicator and cumulative magnitude $|\Delta \tilde{z}|$ and mean axial rate.
    \item \textbf{Visual integrity cues:} quantile summaries of focus $F(t)$,
    contrast $C(t)$, and contamination score $S_{\mathrm{cont}}(t)$.
    \item \textbf{Camera-response descriptors:} signed displacements $(\Delta u_k,\Delta v_k,\Delta z_k)$.
\end{itemize}

To account for inter-patient anatomical variability and differing surgical
styles, all continuous attributes undergo hierarchical normalization. We first
apply within-video z-score normalization to align signals relative to the
procedure-specific baseline. This is followed by a dataset-level robust
scaling (using the $5^{th}$ and $95^{th}$ percentiles) to suppress the influence
of outlier artifacts. This unified, normalized representation allows all events
to be embedded as nodes in an attributed event graph, where temporal overlaps
are preserved as concurrent nodes linked by temporal adjacency edges.

\subsection{Attributed Graph Construction}
\label{sec:graph_attr_matrix}

After extracting event intervals and their multi-modal descriptors, we organize all detected
instances into an \textit{attributed event graph} represented in matrix form by the triplet
$(A, X, S)$. Here, $A$ encodes temporal relations (workflow continuity and concurrency),
$X$ stacks normalized event descriptors, and $S$ captures attribute-based semantic affinity.
This representation enables heterogeneous event categories to be analyzed jointly and supports
downstream strategy mining via WSBGC.

\subsubsection{Temporal Adjacency Matrix}

Let $\{e_k\}_{k=1}^{M}$ denote all detected events defined in
Section~\ref{sec:event_representation}. To preserve both temporal progression and co-occurring
constraints, we connect events that are either sequentially adjacent or temporally overlapping.
We define a binary temporal adjacency matrix $A \in \{0,1\}^{M \times M}$ as
\begin{equation}
A_{ij} = \mathbb{I}\bigl( \text{Seq}(i,j) \ \lor \ \text{Ovl}(i,j) \bigr),
\label{eq:A_def}
\end{equation}
where
\begin{equation}
\text{Seq}(i,j) := \bigl( 0 \le t_j^{s} - t_i^{e} \le \Delta_t \bigr),
\label{eq:Seq_def}
\end{equation}
\begin{equation}
\text{Ovl}(i,j) := \bigl( \min(t_i^{e}, t_j^{e}) - \max(t_i^{s}, t_j^{s}) \bigr) \ge \delta_{\mathrm{ovl}} .
\label{eq:Ovl_def}
\end{equation}
Here, $\Delta_t$ is a small temporal window connecting near-future events, and
$\delta_{\mathrm{ovl}}$ is the minimum overlap duration to be considered concurrent.
The temporal adjacency window was set to $\Delta_t=0.5$\,s, connecting immediately successive or near-concurrent events while preventing excessive graph densification.
We symmetrize the temporal backbone:
\begin{equation}
A \leftarrow \max(A, A^{\top}).
\label{eq:A_sym}
\end{equation}

\subsubsection{Node Attribute Matrix}

Stacking the event descriptors yields the attribute matrix
\begin{equation}
X =
\begin{bmatrix}
\mathbf{x}_1^{\top} \\
\mathbf{x}_2^{\top} \\
\vdots \\
\mathbf{x}_M^{\top}
\end{bmatrix}
\in \mathbb{R}^{M \times d},
\label{eq:X_def}
\end{equation}
where each $\mathbf{x}_k$ is the normalized state--action descriptor defined in
Section~\ref{sec:event_representation}, with
non-applicable dimensions masked per event type.

Normalization follows the procedure described in
Section~\ref{sec:event_representation}, followed by row-wise $\ell_2$ normalization:
\begin{equation}
\hat{\mathbf{x}}_k = \frac{\mathbf{x}_k}{\|\mathbf{x}_k\|_2 + \epsilon}.
\label{eq:l2_norm}
\end{equation}
For simplicity, $\mathbf{x}_k$ denotes the normalized descriptor hereafter.

\subsubsection{Attribute-Based Similarity Matrix}

Temporal adjacency captures workflow structure but does not reveal semantic relations between
events that share similar descriptors despite being temporally distant. We therefore compute an
attribute-based similarity matrix $S$ from the normalized descriptors.

Because descriptors contain masked dimensions, cosine similarity is computed only over valid
(unmasked) feature dimensions. Let $\mathbf{m}_k \in \{0,1\}^{d}$ denote the validity mask of
event $e_k$, and $\mathbf{m}_{ij}=\mathbf{m}_i \wedge \mathbf{m}_j$ be the pairwise valid mask.
We define masked cosine similarity as
\begin{equation}
S_{ij} =
\frac{
\bigl( (\mathbf{m}_{ij} \odot \mathbf{x}_i)^{\top} (\mathbf{m}_{ij} \odot \mathbf{x}_j) \bigr)
}{
\| \mathbf{m}_{ij} \odot \mathbf{x}_i \|_2 \, \| \mathbf{m}_{ij} \odot \mathbf{x}_j \|_2 + \epsilon
},
\label{eq:masked_cosine}
\end{equation}
and we set $S_{ij}=0$ when $\|\mathbf{m}_{ij}\|_1$ is below a small threshold.

To reduce noise and avoid overly dense connectivity, we retain only the top-$k$ most similar
neighbors per event. 
The attribute similarity graph retained the top-$k$ neighbors with $k=20$, which maintains sufficient relational connectivity without introducing spurious long-range links.
\begin{equation}
S_{ij} \leftarrow
\begin{cases}
S_{ij}, & \text{if } j \in \mathrm{TopK}(i),\\
0,      & \text{otherwise},
\end{cases}
\label{eq:topk}
\end{equation}
followed by symmetrization:
\begin{equation}
S \leftarrow \max(S, S^{\top}).
\label{eq:S_sym}
\end{equation}

The triplet $(A, X, S)$ defines the attributed event graph used by the downstream WSBGC stage:
$A$ provides a temporal backbone (sequential and concurrent relations), while $S$ captures
attribute-level proximity in the normalized descriptor space.

\subsection{Strategy Mining via Graph Clustering}
\label{sec:wsbgc}

\begin{figure*}[htbp]
    \centering
    \includegraphics[width=0.95\textwidth, trim=0 90 0 80, clip]{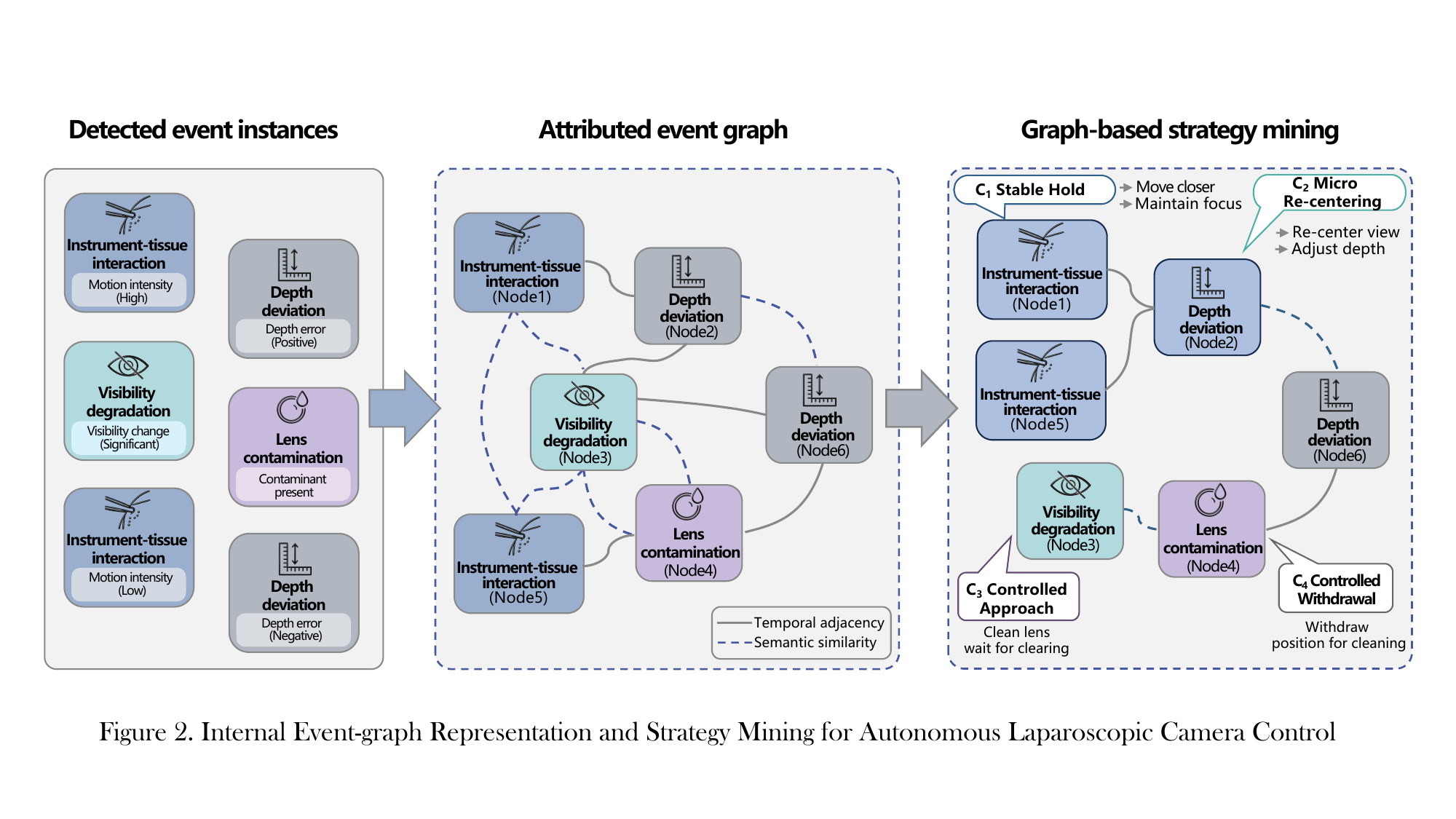}
    \caption{\textbf{Event graph representation and strategy mining:} Camera-relevant events are represented as nodes in an attributed graph, with temporal and semantic relations encoded as heterogeneous edges. WSBGC groups events into coherent clusters corresponding to reusable camera-handling strategies.}
    \label{fig:event_graph}
\end{figure*}

Section~\ref{sec:offline_event_parsing} defines a compact event taxonomy for surgical
situations and augments each detected interval with a multi-modal descriptor $\mathbf{x}_k$,
which includes the camera adjustment observed over the same interval. While event types
describe \emph{what} is happening, they do not uniquely specify \emph{how} the laparoscope is
handled: similar situations can be accompanied by different adjustments (e.g., re-center left
vs.\ right, stabilize vs.\ compensate, or varying depth-correction magnitudes), and similar
adjustments can recur under different situational contexts. We therefore cluster events to
identify recurrent \emph{camera-response patterns} (strategy primitives) that emerge beyond
coarse event labels.

Given the triplet $(A, X, S)$ from Section~\ref{sec:graph_attr_matrix}, we perform graph-boosted
clustering following the WSBGC formulation (Section~\ref{sec:rw_wsbgc}). The temporal backbone
$A$ preserves workflow continuity and co-occurrence, while the affinity $S$ links events that
are similar in descriptor space. WSBGC refines these relations, learns node embeddings, and
produces clusters that are subsequently interpreted and stored in the strategy knowledge base
(Section~\ref{sec:strategy_kb}).

\subsubsection{Graph Boosting and Refined Graph Construction}

We fuse $A$ and $S$ to form an auxiliary boosted graph $\mathbf{G}_A$ via Eq.~\eqref{eq:GA},
and apply the self-expressiveness refinement in Eq.~\eqref{eq:selfexpress} to obtain
$\mathbf{S}_c$ (Eq.~\eqref{eq:Sc}). The refined graph $\mathbf{G}_R$ is then constructed via
Eq.~\eqref{eq:GR}, which enhances reliable connectivity while suppressing spurious edges under
heterogeneous surgical conditions.

\subsubsection{Node Embedding via Graph Attention Autoencoder}

Although $\mathbf{G}_R$ provides enriched connectivity, directly clustering on adjacency entries
can be sensitive to residual noise and uneven density. We therefore learn compact node embeddings
using a graph attention autoencoder (GAAE). The encoder performs attention-weighted message
passing over $\mathbf{G}_R$:
\begin{equation}
    \mathbf{z}_i
    =
    \sigma
    \left(
        \sum_{j \in \mathcal{N}(i)}
        \alpha_{ij}\,
        \mathbf{W}_{e}\mathbf{x}_j
    \right),
    \label{eq:gaae_enc}
\end{equation}
where $\mathbf{x}_j$ is the attribute vector of node $j$, $\mathbf{W}_{e}$ is a learnable
projection matrix, and $\sigma(\cdot)$ is a nonlinear activation. where $\alpha_{ij}$ denotes the attention weight learned over the neighborhood of $\mathcal{N}(i)$.

The decoder reconstructs the refined graph:
\begin{equation}
    \hat{\mathbf{G}}_{R}
    =
    \sigma\!\left(
        \mathbf{Z}\mathbf{Z}^{\top}
    \right),
    \label{eq:gaae_dec}
\end{equation}
where $\mathbf{Z}=[\mathbf{z}_1^{\top},\ldots,\mathbf{z}_M^{\top}]^{\top}$.
Training minimizes a weighted reconstruction loss:
\begin{equation}
    \mathcal{L}_{\mathrm{rec}}
    =
    \left\|
        \mathbf{W}
        \odot
        \left(
            \mathbf{G}_{R}
            -
            \hat{\mathbf{G}}_{R}
        \right)
    \right\|_F^2,
    \label{eq:gaae_loss}
\end{equation}
where $\mathbf{W}$ is the confidence weighting used in Eq.~\eqref{eq:WSNMF}, emphasizing reliable
relations while downweighting uncertain connections.

\subsubsection{Weighted Symmetric NMF Clustering}

We extract strategy clusters by applying weighted symmetric NMF on $\mathbf{G}_R$
using Eq.~\eqref{eq:WSNMF}, yielding a non-negative cluster-indicator matrix
$\mathbf{U} \in \mathbb{R}^{M \times K}$, where each row encodes soft memberships
over $K$ clusters. Each cluster represents a recurrent camera-handling primitive
characterized by consistent event context and motion-response patterns in the
descriptor space.

In the primary configuration, we set $K=12$ and assign a single dominant
cluster label $C_K$ to each time step based on the maximum membership in
$\mathbf{U}$. The discovered 12 strategy primitives are summarized
in Table~\ref{tab:strategy_clusters}. The sensitivity of the framework to
the choice of $K$ is further analyzed in the experimental section.

\begin{table}[t]
\caption{Summary of the 12 mined strategy clusters.}
\label{tab:strategy_clusters}
\centering
\setlength{\tabcolsep}{3.5pt}
\begin{tabular}{cl}
\toprule
\textbf{Cluster} & \textbf{Strategy Description} \\
\midrule
$C_1$  & Stable hold with neutral directions. \\
$C_2$  & Small translation for tool re-centering. \\
$C_3$  & Depth-dominant controlled approach. \\
$C_4$  & Depth-dominant controlled withdrawal. \\
$C_5$  & Horizontal-dominant motion tracking. \\
$C_6$  & Vertical-dominant motion tracking. \\
$C_7$  & Yaw/pitch-based view reframing. \\
$C_8$  & Roll-based view leveling. \\
$C_9$  & Visibility-driven mild retreat and reframing. \\
$C_{10}$ & Contamination-triggered withdrawal. \\
$C_{11}$ & Local workspace shift with small composite motion. \\
$C_{12}$ & Global workspace transition with larger motion. \\
\bottomrule
\end{tabular}
\end{table}

\subsection{Strategy Primitive Extraction and Supervision Construction}
\label{sec:strategy_kb}

The WSNMF clustering partitions events into
$K$ groups $\{C_k\}_{k=1}^{K}$ that share consistent descriptor statistics
and relational structure. These clusters represent recurrent
\emph{camera-handling primitives}, which we formalize as strategy labels
used for supervision.

\subsubsection{Cluster-Level Strategy Semantics}

For each cluster $C_k$, we compute a robust prototype using the
normalized event descriptors:
\begin{equation}
\bar{\mathbf{x}}_k
=
\mathrm{median}_{e_i \in C_k} (\mathbf{x}_i),
\end{equation}
where $\mathbf{x}_i$ is defined in
Section~\ref{sec:event_representation}. The dominant event type is
determined by majority composition over taxonomy labels.

Each cluster is then assigned a strategy name according to
deterministic rules derived from its prototype statistics
(e.g., signed depth displacement, re-centering direction,
visibility indicators). The resulting strategy set
$\mathcal{S} = \{s_k\}_{k=1}^{K}$ forms a compact library of
recurrent camera-response modes.

\subsubsection{Directional Action Label Construction}

While strategy labels describe recurrent response patterns,
the controller requires explicit motion supervision in the same command space
used by the RCM-constrained IBVS controller.

For each event interval $e_i = (y_i, t_i^s, t_i^e, \mathbf{x}_i)$,
we directly use the interval-level motion descriptor
$\mathbf{x}^{\mathrm{action}}_i = (\Delta u_i,\Delta v_i,\Delta z_i)$
to construct the directional supervision signal.

To obtain robust labels suitable for closed-loop control, we discretize
$\mathbf{x}^{\mathrm{action}}_i$ into a 3-DoF directional vector:
\begin{equation}
\mathbf{d}_i \in \{-1,0,+1\}^{3},
\end{equation}
where each component corresponds to the signed direction along
$(u,v,z)$. Motion magnitudes below fixed deadband thresholds are
mapped to zero (\emph{hold}) to suppress estimation noise.

\subsubsection{Supervision Dataset Construction}

Training samples are constructed at the event level.
For each detected interval $[t_i^s, t_i^e]$, we form:
\begin{equation}
\mathcal{D}_i
=
\left(
\mathbf{V}_{t_i^s:t_i^e},
\; s_{k_i},
\; \mathbf{d}_i
\right),
\end{equation}
where $\mathbf{V}_{t_i^s:t_i^e}$ denotes a short video segment covering the event.

The strategy label $s_{k_i}$ provides an interpretable
intermediate supervisory signal, while the directional vector
$\mathbf{d}_i$ supervises camera motion prediction.
This representation decouples high-level response patterns
from metric motion magnitudes and is well suited for
closed-loop visual servo control.

At inference time, optional speech input is integrated
as an external override channel that biases or replaces
the predicted direction, without requiring speech data
during training.

\subsection{Real-Time Strategy-Supervised Direction Prediction}
\label{sec:rt_direction}

The direction labels encode sign only; motion magnitude is regulated by the
IBVS--RCM controller (Appendix~B), separating high-level intent prediction from
low-level execution.

Although labels are defined at the event-interval level, training is performed
at the frame level. For any frame $t$ within an event interval
$[t_i^s,t_i^e]$, we assign the supervision
\begin{equation}
(s_t^\ast,\mathbf{d}_t^\ast) = (C_i,\mathbf{d}_i),
\end{equation}
which links each frame to a temporally coherent camera-response segment and
suppresses frame-wise noise.

Let $\mathbf{I}_t$ denote the laparoscopic image at time $t$.
A visual encoder $\Phi(\cdot)$ produces a feature embedding
\begin{equation}
\mathbf{h}_t = \Phi(\mathbf{I}_t), \qquad \mathbf{h}_t \in \mathbb{R}^{D}.
\end{equation}
Two prediction heads are attached to $\mathbf{h}_t$.

\paragraph{Strategy Head.}
The strategy head predicts the offline-discovered strategy primitive:
\begin{equation}
\hat{\mathbf{s}}_t = \mathrm{Softmax}(\mathbf{W}_s \mathbf{h}_t),
\end{equation}
where $\hat{\mathbf{s}}_t \in \mathbb{R}^{K}$.

\paragraph{Direction Head.}
The direction head predicts a 3-DoF discrete command in the image-based
command space:
\begin{equation}
\hat{\mathbf{p}}^{(j)}_t
=
\mathrm{Softmax}(\mathbf{W}^{(j)}_d \mathbf{h}_t),
\qquad j\in\{u,v,z\},
\end{equation}
where $\hat{\mathbf{p}}^{(j)}_t \in \mathbb{R}^{3}$ corresponds to
$\{-1,0,+1\}$. The predicted direction is
\begin{equation}
\hat{d}^{(j)}_t = \arg\max \hat{\mathbf{p}}^{(j)}_t,
\end{equation}
and the full command is
\begin{equation}
\hat{\mathbf{d}}_t = (\hat{d}^{(u)}_t,\hat{d}^{(v)}_t,\hat{d}^{(z)}_t)
\in \{-1,0,+1\}^{3}.
\end{equation}

\paragraph{Training Objective.}
Given supervision $(s_t^\ast,\mathbf{d}_t^\ast)$, the loss combines directional
classification with auxiliary strategy supervision:
\begin{equation}
\label{loss}
\mathcal{L}
=
\mathcal{L}_{\mathrm{dir}}
+
\lambda_s \mathcal{L}_{\mathrm{str}},
\end{equation}
with
\begin{equation}
\mathcal{L}_{\mathrm{dir}}
=
\frac{1}{3}
\sum_{j\in\{u,v,z\}}
\mathrm{CE}\!\left(d_{t}^{\ast(j)}, \hat{\mathbf{p}}^{(j)}_t\right),
\end{equation}
and
\begin{equation}
\mathcal{L}_{\mathrm{str}}
=
\mathrm{CE}\!\left(s_t^\ast, \hat{\mathbf{s}}_t\right).
\end{equation}
The auxiliary strategy loss preserves the cluster-level structure discovered
offline and improves robustness under visually ambiguous conditions.

\paragraph{Closed-Loop Deployment.}
During deployment, only $\hat{\mathbf{d}}_t$ is forwarded to the control layer.
Following Appendix~B, $(\hat{d}^{(u)}_t,\hat{d}^{(v)}_t)$ determines the sign of
the setpoint shift in the image plane, while $\hat{d}^{(z)}_t$ updates the
penetration setpoint. The IBVS--RCM controller computes the resulting motion
magnitude and enforces RCM and safety constraints in closed loop:
\[
\mathbf{I}_t
\;\rightarrow\;
\hat{\mathbf{d}}_t
\;\rightarrow\;
\text{IBVS--RCM control}
\;\rightarrow\;
\mathbf{I}_{t+1}.
\]

\section{Experiments}

\subsection{Implementation Details}

Our dataset consists of 109 laparoscopic cholecystectomy cases, including 50 private expert recordings and 59 high-quality camera-control videos selected from Cholec80. All videos are resized to $512\times512$ before processing.

Training samples are extracted only within valid camera-adjustment intervals detected by the offline event parsing module. Frames inside these intervals are sampled at 10\,Hz, matching the intended online decision update rate. Each case yields 10,800 labeled samples, resulting in a total of 1,177,200 raw samples.

To address the dominance of neutral (hold) directions, all non-zero direction samples are retained while zero-direction samples are randomly downsampled to match the total number of non-zero samples. After balancing, the final dataset contains 706,320 supervised samples.

The dataset is split into 600,000 training samples, 53,160 validation samples, and 53,160 test samples.

We adopt Qwen2.5-VL 7B as the backbone and attach a strategy head and a direction head. The training objective follows Eq.~\ref{loss}.

Low-Rank Adaptation (LoRA) is applied to the query and value projection matrices of all transformer layers in the language backbone. The LoRA rank is set to $r=16$, with scaling factor $\alpha=32$ and dropout 0.05.

Training is conducted using mixed precision (bfloat16) on 4 $\times$ NVIDIA A100 80GB GPUs. The effective batch size is 64 (per-device batch size 8 with gradient accumulation). The learning rate is $2\times10^{-4}$ with cosine decay and 5\% warm-up. We train for 3 epochs, corresponding to approximately 28,000 optimization steps. Total training time is approximately 22 hours.

\subsection{Experimental Setup}

\begin{figure*}[htbp]
    \centering
    \includegraphics[width=0.9\linewidth]{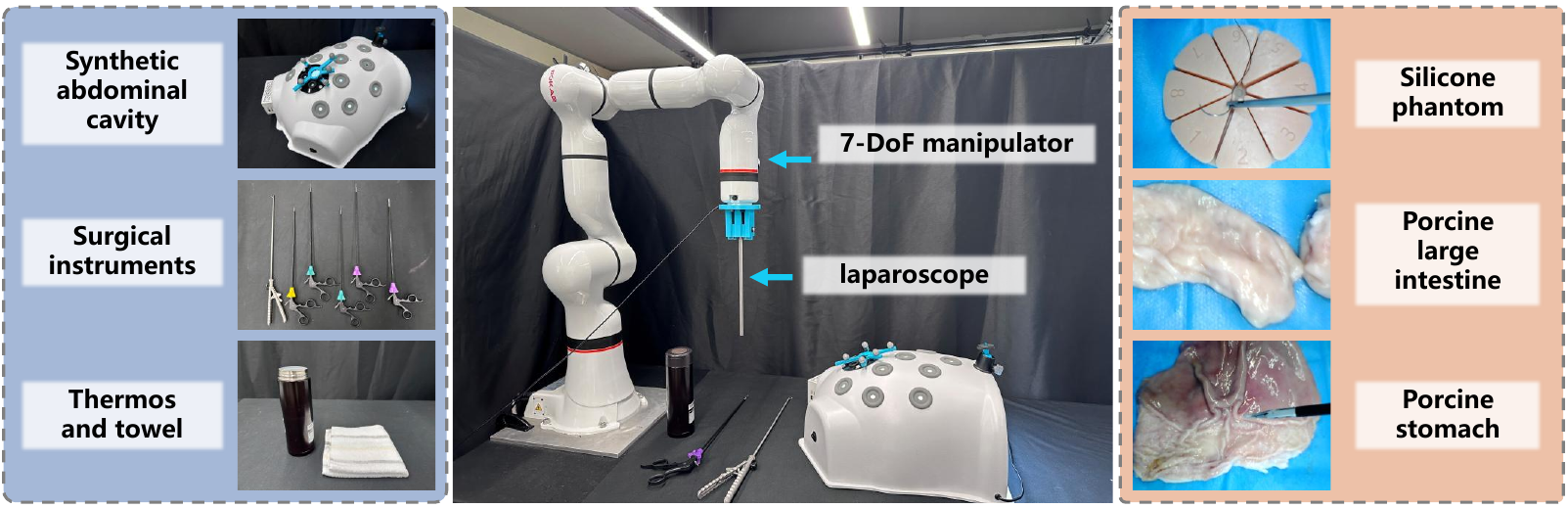}
    \caption{\textbf{Experimental setup:} From left to right, the surgical equipments used in the experiments, including the synthetic abdominal cavity, surgical instruments, and anti-fog devices; an overview of the surgical workspace with the robotic manipulator; and the manipulated objects, consisting of a silicone phantom and porcine tissues used to mimic real organ.}
    \label{surgical-experimental-setup}
\end{figure*}

All experiments were conducted using a robotic laparoscopic camera system, as shown in Fig.~\ref{surgical-experimental-setup}. A standard rigid laparoscope is rigidly mounted on a ROKAE ER7Pro robotic manipulator, providing a monocular RGB stream at 30~fps with a resolution of $640\times480$.

A synthetic abdominal cavity is used to emulate realistic operative conditions and approximate the anatomical constraints of the patient body. Experimental targets include a soft silicone phantom and ex vivo porcine large intestine and stomach tissues, on which stitching and membrane dissection tasks are performed. Surgical instruments (needle holder, curved dissecting forceps, double-action curved scissors, and fine-tooth tissue forceps) are manually manipulated inside the cavity, producing representative camera-relevant motion patterns, including deliberate working-distance variations, rapid instrument sweeps, and fine dissection-like movements. These demonstrations are used both for offline strategy extraction and for evaluating the proposed real-time autonomous laparoscopic camera control framework.

The full perception--inference--control pipeline runs on a single workstation equipped with an NVIDIA GeForce RTX 5090 GPU. During deployment, the multi-modal VLM updates discrete motion directions asynchronously at 10~Hz, while the IBVS--RCM controller executes continuous visual servoing at 100~Hz. The VLM is quantized to 4-bit weights using AWQ for efficient inference. When speech input is enabled, it is captured via an onboard microphone and processed online by a lightweight speech encoder in real time. The overall system, including sensing, inference, safety filtering, data transfer, and motion execution, is synchronized to ensure consistent closed-loop performance.

This hardware configuration enables controlled and repeatable experiments, supporting systematic evaluation of both the accuracy of the learned camera directives and the stability of the proposed autonomous laparoscopic control framework.

\subsection{Event Detection and Strategy Clustering Evaluation}
\label{sec:event-cluster-eval}

To validate the reliability of the offline perception and strategy-mining
pipeline, we evaluate (i) the accuracy of the event detection modules and
(ii) the semantic consistency and structural compactness of the mined strategy
clusters. In all primary experiments, we use $K=12$ strategy clusters, which
define the dominant strategy labels used for downstream supervision and online
control. We additionally design an ablation protocol to assess the sensitivity
to the choice of $K$.

Since event-level evaluation requires temporally precise ground-truth
boundaries, we construct a dedicated annotation subset by stratified sampling
across different videos and candidate strategy clusters to ensure broad
coverage of camera behaviors. From held-out videos, we select 162 camera-active
temporal segments (totaling 94 minutes) with frequent viewpoint adjustments.
Three surgeons independently annotate event intervals using start--end
timestamps according to the event taxonomy defined in Section~IV. These
annotated sequences are used for both event detection evaluation and strategy
cluster analysis.

\subsubsection{Event Detection Evaluation}

We evaluate event localization at the segment level. A predicted event is
counted as a true positive if it matches the ground-truth event type and the
temporal Intersection-over-Union (tIoU) between predicted and annotated
intervals is at least 0.5. Based on this matching criterion, we report
class-wise Precision, Recall, and F1-score, together with the mean tIoU over
matched detections. For depth deviation events, we additionally report the
mean absolute error (MAE) of the estimated distance change to the working
surface over the annotated interval. These metrics jointly assess event-type
correctness, temporal alignment, and quantitative accuracy of the detection
pipeline.

\subsubsection{Strategy Cluster Evaluation}

To assess whether the mined clusters correspond to meaningful camera-handling
strategies, we perform a semantic consistency analysis on the same held-out
annotation subset. For each cluster, we select 30 representative event clips
nearest to the cluster centroid in the embedding space. Three surgeons
independently assign each clip to one of the predefined strategy categories
(e.g., working-distance maintenance, field-of-view re-centering, visibility
recovery, contamination handling). We report clustering Purity and Normalized
Mutual Information (NMI) between cluster assignments and surgeon-provided
semantic labels to quantify semantic alignment.

To further evaluate structural compactness, we compute the intra-cluster
variance of strategy-relevant attributes. For each event $i$, we define the
attribute vector
$\mathbf{a}_i = [d_i,\, f_i]^{\top}$,
where $d_i$ denotes depth deviation and $f_i$ captures optical-flow
directionality. For each cluster $C_k$, the intra-cluster variance is defined as
\begin{equation}
    \mathrm{Var}_{\mathrm{intra}}(C_k)
    =
    \frac{1}{|C_k|}
    \sum_{i \in C_k}
    \left\|
        \mathbf{a}_i - \bar{\mathbf{a}}_k
    \right\|_2^{2},
\end{equation}
where $\bar{\mathbf{a}}_k$ denotes the mean attribute vector of cluster $C_k$.
Lower intra-cluster variance indicates that events grouped within the same
cluster exhibit consistent depth profiles and motion characteristics,
reflecting a coherent and distinctive camera-handling strategy.

\subsubsection{Ablation Protocol on the Number of Clusters $K$}
\label{sec:ablation_k_protocol}

To examine the sensitivity of strategy mining to the number of clusters,
we evaluate multiple values of $K$ around the default setting.
Specifically, we consider $K \in \{8, 10, 12, 14, 16\}$,
spanning coarse-to-fine strategy granularity.
For each $K$, the attributed event graph is re-clustered under identical
settings, and clustering quality is evaluated using Purity, Normalized
Mutual Information (NMI), and the average intra-cluster variance
$\mathrm{Var}_{\mathrm{intra}}$. These metrics quantify semantic consistency
with expert annotations and structural compactness in the attribute space.
This protocol allows systematic assessment of how cluster granularity
affects the coherence and interpretability of the discovered strategy
primitives. 

\subsection{Ex Vivo Laparoscopic Camera-Control Evaluation}

To evaluate the effectiveness of the proposed autonomous laparoscopic camera controller under realistic operative conditions, ex vivo experiments were conducted using a silicone phantom, porcine large intestine, and porcine stomach tissue. The experimental tasks consisted of standardized stitching and membrane dissection procedures, during which the surgeons manipulated various surgical instruments while the laparoscopic camera was controlled either manually or by the proposed system.

This evaluation quantitatively assesses the ability of the autonomous controller to maintain an optimal surgical field of view, appropriate working distance, and stable camera pose throughout dynamic tissue manipulation. Additionally, the responsiveness of the VLM is also assessed.

\subsubsection{Experimental Protocol}
\label{experimental-protocol}

During the experiments, a silicone phantom, porcine large intestines, and porcine stomach tissues was placed respectively inside a synthetic abdominal cavity to recreate the confined workspace and visual constraints characteristic of MIS. The trocar was positioned at a fixed entry point, and a laparoscopic camera mounted on the robotic manipulator was inserted through the trocar. The surgeon then performed stitching and membrane dissection tasks along adjacent tissue regions and within a predefined area of the gastric wall using standard surgical instruments. During the procedure, the tissues underwent traction, rotation, and intermittent occlusions, thereby creating challenging conditions for maintaining a stable and informative surgical view. Trials were conducted under two different camera control conditions:

\begin{itemize}
    \item \textbf{Manual camera control (Human).} A junior surgeon who satisfies institutional laparoscopic camera-handling training standards manually manipulates the laparoscope to maintain an appropriate surgical view throughout the procedure.
    
    \item \textbf{Autonomous control with speech refinement.} The laparoscope is controlled autonomously by the manipulator. Moreover, in addition to autonomous control, the surgeon can provide brief verbal commands (e.g., ``closer'', ``upward'') through the VLM interface to further refine the selected viewpoint.
\end{itemize}


For each control condition, a total of $3$ silicone phantom stitching trials, $3$ porcine large intestine stitching trials, and $3$ porcine stomach membrane dissection trials were performed by surgeons of $3$ different experience levels (senior, junior, and intern) to account for inter-operator variability in manipulation style. Each trial lasted approximately $2$--$14$ minutes, depending on the surgeon’s proficiency and the complexity of the task. All experiments were video-recorded and synchronized with the corresponding robot state logs, enabling comprehensive quantitative evaluation of the proposed system.

\subsubsection{Evaluation Metrics}

Performance is evaluated through a combination of objective quantitative metrics.
For the quantitative assessment, the following metrics are utilized to evaluate the performance of the proposed laparoscopic control method and compare it against the one of the conventional manual camera control:

\textbf{Field-of-View Centering:} The consistency of maintaining the operation region near the desired pixel coordinate is quantified as follows: Let $\vec{f}\left(t\right) \in \mathbb{R}^2$ denote the current tip pixel coordinate at time $t \in \mathbb{R}$, and $\vec{f}_{d}\left(t\right) \in \mathbb{R}^2$ denotes the desired pixel coordinate at time $t$. The centering error $e_{cen} \in \mathbb{R}$ is defined as:
\begin{equation}
    e_{cen}
    =
    \frac{1}{T}\sum_{t=1}^{T}
    \left\|
        \vec{f}\left(t\right) - \vec{f}_{d}\left(t\right)
    \right\|_2.
\end{equation}
Here, $T \in \mathbb{R}$ denotes the trial duration. Lower values indicate improved view maintenance.


\textbf{Image Shaking:} View instability is quantified by detecting salient features in the first valid frame and tracking them throughout the image sequence using the KLT optical flow tracker. For each pair of consecutive frames, the inter-frame displacement of successfully tracked features is computed as a robust estimate of camera-induced image motion. Let the feature point set at time $t$ be denoted as $\vec{x}_{k}\left(t\right) := \left[u_{k}\left(t\right),v_{k}\left(t\right)\right]^\top \in \mathbb{R}^2$, in which $u_{k}\left(t\right)$ and $v_{k}\left(t\right)$ denote the pixel coordinate of the selected feature. The displacement of each feature between two adjacent frames is $\Delta \vec{x}_{k}\left(t\right) := \vec{x}_{k}\left(t\right)-\vec{x}_{k}\left(t-1\right) \in \mathbb{R}^2$, and the frame-wise shaking magnitude is defined as the median value of the feature displacements: $s\left(t\right) = \mathrm{median}\left(\|\Delta \vec{x}_{k}\left(t\right)\|_2\right)$. The overall image shaking index $S \in \mathbb{R}$ is then obtained by averaging this quantity over the entire trial duration:
\begin{equation}
    S 
    = 
    \frac{1}{T}\sum_{t=1}^{T}
    s\left(t\right).
\end{equation}
Lower values of $S$ indicate reduced image shaking and improved view stability.
    
\textbf{Time to Target:} The elapsed time interval $\delta t_{target} \in \mathbb{R}$ required for the center of the laparoscopic camera FoV to reach the desired target location from its current position is measured.



In addition to the aforementioned metrics, the following measures are specifically introduced to further assess the performance and effectiveness of the proposed methods:

\begin{figure*}[htbp]
    \centering
    \includegraphics[width=0.9\linewidth]{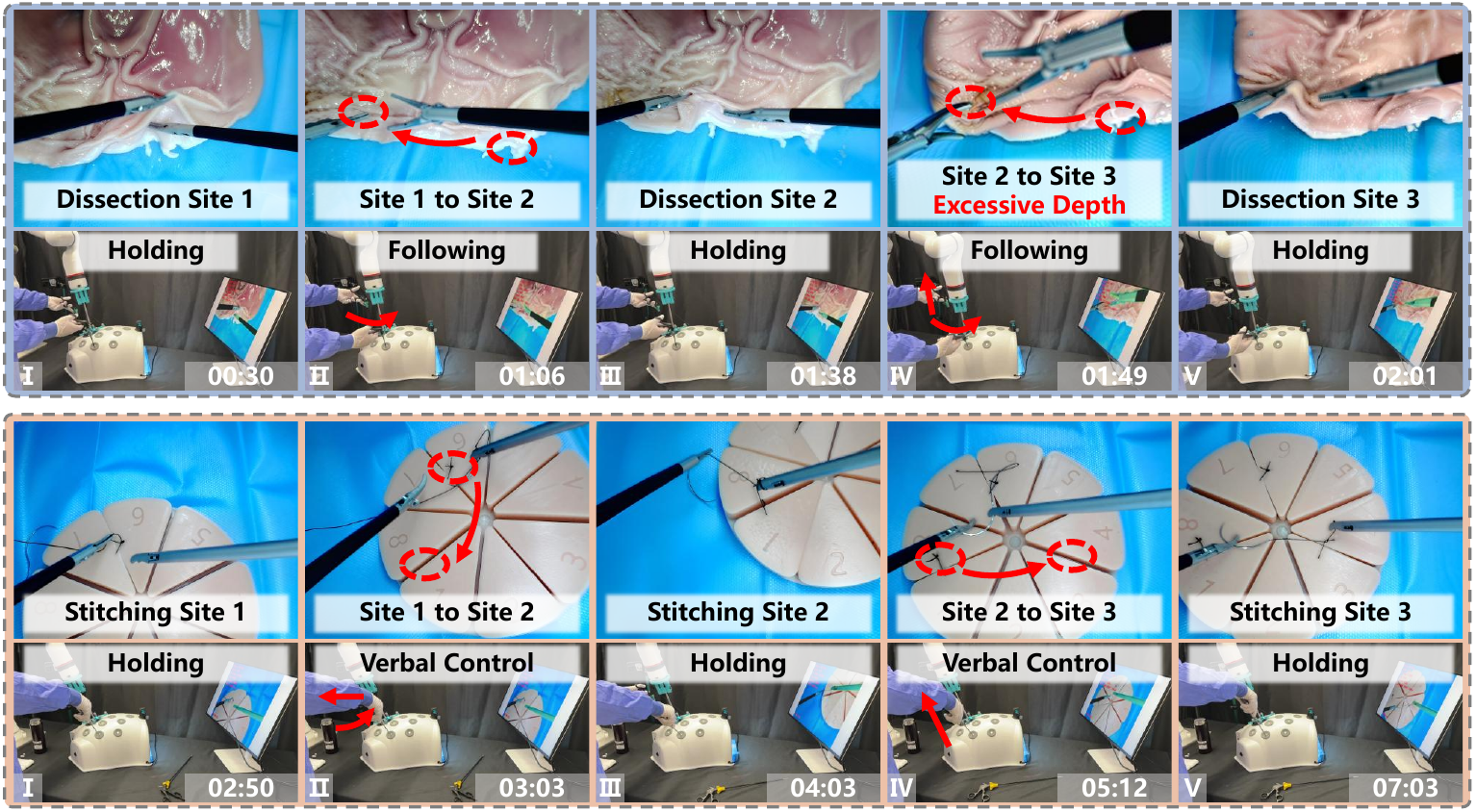}
    \caption{\textbf{Experimental validation of the proposed autonomous laparoscopic control method in complex surgical tasks: } Upper part: Illustration of a multi-site dissection task performed on porcine gastric tissue. The images demonstrate the autonomous following behavior of the laparoscope, as well as its ability to adapt the penetration ratio in response to shifts in the surgical instrument’s workspace. Lower part: Validation during a simulated silicone suturing task. The system exhibits multi-modal interaction capability, allowing the surgeon to refine the laparoscope pose via voice commands without interrupting the manual suturing process. The timestamps shown in the bottom-right corner highlight the continuous operation and efficiency of the proposed robotic assistance system.}
    \label{surgical-validation}
\end{figure*}

\begin{figure}[htbp] 
    \centering
    \includegraphics[width=0.9\columnwidth]{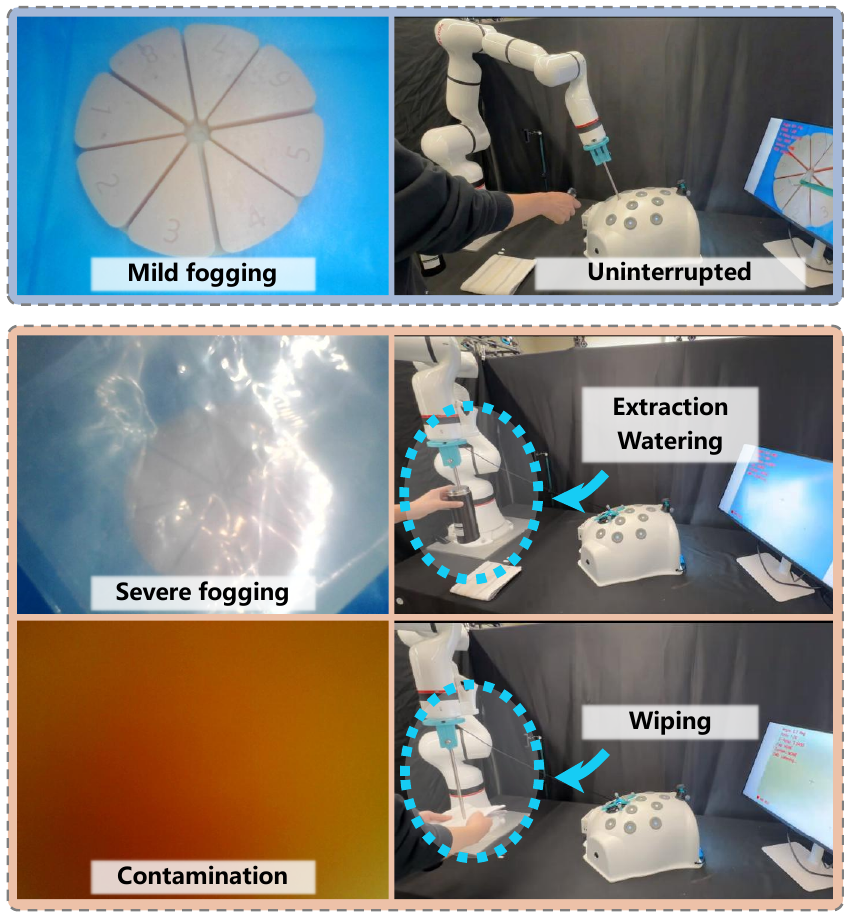}
    \caption{\textbf{Experimental validation of the VLM-based fogging and contamination detection capability:} Upper part: System response under mild fogging conditions, where the laparoscope remains stationary, enabling uninterrupted surgical operation. Lower part: Autonomous detection and recovery under severe fogging and contamination. Upon detecting severe fogging, the system automatically initiates an ``extraction–watering–wiping'' sequence to restore visual clarity. On the other hand, once contamination (e.g., blood or tissue debris) is located on the lens, the system triggers the same cleaning procedure as in the severe fogging scenario to ensure a clear surgical view. The results demonstrate the robustness of the proposed framework in maintaining an optimal surgical field of view (FoV) through autonomous instrument withdrawal and lens cleaning.}
    \label{fig:autonomous-laparoscopic-clean-validation}
\end{figure}

\begin{figure}[t]
    \centering
    \includegraphics[width=0.95\linewidth]{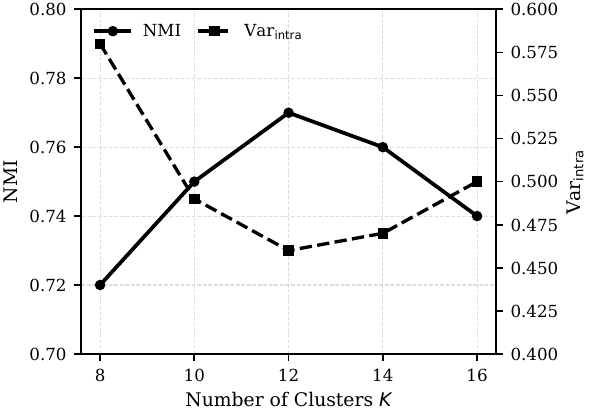}
    \caption{\textbf{Effect of the number of clusters $K$ on clustering quality:} As $K$ increases from 8 to 12, NMI improves and intra-cluster variance decreases, indicating more coherent strategy grouping. For $K \ge 14$, gains saturate and mild over-segmentation effects emerge.}
    \label{fig:k_ablation}
\end{figure}

\textbf{Working Distance Stability:} The stability of the working distance is assessed by computing the absolute and relative errors, denoted as $e_{depth} \in \mathbb{R}$ and $\hat{e}_{depth} \in \mathbb{R}$, between the current and desired penetration ratio of the RCM point along the laparoscope shaft:
\begin{align}
    e_{depth}
    &=
    \frac{1}{T}\sum_{t=1}^{T}
    \left| \lambda\left(t\right) - \lambda_{d}\left(t\right) \right|, \\[6pt]
    \hat{e}_{depth}
    &=
    \frac{1}{T}\sum_{t=1}^{T}
    \dfrac{\left| \lambda\left(t\right) - \lambda_{d}\left(t\right) \right|}{\lambda_{d}\left(t\right) + \epsilon}.
\end{align}
Here, $\epsilon \in \mathbb{R}$ is a small positive number. Larger errors indicate inferior zoom regulation and potential misalignment with respect to the surgical target, whereas consistently small error reflect stable depth control.

\textbf{High-Frequency Ratio:} The discrete Fourier transform (DFT) is applied to the positional trajectory of the instrument tip to perform frequency-domain analysis. The power spectral density along each of the three axes is computed, and the proportion of spectral energy at frequencies greater than or equal to $4$ Hz relative to the total spectral energy is calculated for each axis. The resulting high-frequency energy ratio is defined as $\vec{R} := \left[R_x, R_y, R_z\right]^\top \in \mathbb{R}^3$, which quantifies the high-frequency vibration components of the trajectory:
\begin{equation}
    M_i(k) 
    = 
    \sum_{n=0}^{N-1} m_i(n)\,
    e^{-j 2\pi kn / N},
    \quad i \in \{x,y,z\},
\end{equation}
\begin{equation}
    R_i 
    = 
    \frac{
    \sum_{f_k \ge 4} \left| M_i(k) \right|^2
    }{
    \sum_{k=0}^{N-1} \left| M_i(k) \right|^2
    }.     
\end{equation}
Here, $m_i(n)$ denotes the discrete trajectory samples along axis $i$, 
and $M_i(k)$ represents its discrete Fourier transform. $f_k$ is the corresponding frequency of the $k$-th spectral component. $R_i$ quantifies the proportion of high-frequency spectral energy ($\geq 4$ Hz) along axis $i$. A lower value of $\vec{R}$ indicates that the trajectory is predominantly smooth with limited high-frequency oscillatory components.

\textbf{Pose Jitter:} Motion jitter is quantified by applying a low-pass filter to decompose the camera velocity signal $\vec{v}\left(t\right) \in \mathbb{R}^3$ into low- and high-frequency components, denoted as $\vec{v}_{low}\left(t\right) \in \mathbb{R}^3$ and $\vec{v}_{high}\left(t\right) \in \mathbb{R}^3$, respectively. The jitter index $J \in \mathbb{R}$ is then defined as the root-mean-square (RMS) magnitude of the high-frequency component:
\begin{align}
    &\vec{v}_{jitter}\left(t\right) = \vec{v}\left(t\right)-\vec{v}_{low}\left(t\right), \\[6pt]
    &J = \sqrt{\dfrac{1}{T}\sum_{t=1}^{T} \| \vec{v}_{jitter}\left(t\right)\|_2}.
\end{align}
A higher value of $J$ indicates stronger high-frequency oscillations and reduced motion stability.




\textbf{Recovery Time:} The elapsed time $\delta t_{recover} \in \mathbb{R}$ required for the manipulator to retract the laparoscope, execute the cleaning procedure, and subsequently restore the scope to its original pose prior to retraction.

These metrics collectively evaluate whether the proposed system can maintain a stable and informative viewpoint throughout dynamic ex vivo dissection tasks, and whether verbal or visual refinement improve practical usability during operative manipulation.

\section{Results and Discussion}
\subsection{Event Detection and Strategy Clustering Results}

\begin{table}[t]
    \caption{Performance of Event Detection ($\mathrm{tIoU}\ge 0.5$)}
    \label{tab:event_strategy_results}
    \centering
    \setlength{\tabcolsep}{5pt}
    \begin{tabular}{lcccc}
        \toprule
        \textbf{Event} 
        & \textbf{Precision} 
        & \textbf{Recall} 
        & \textbf{F1-score} 
        & \textbf{Mean tIoU} \\
        \midrule
        Interaction 
        & 0.86 & 0.81 & 0.83 & 0.67 \\
        Depth deviation 
        & 0.90 & 0.88 & 0.89 & 0.74 \\
        Visibility degradation 
        & 0.84 & 0.79 & 0.81 & 0.64 \\
        Lens contamination 
        & 0.93 & 0.91 & 0.92 & 0.80 \\
        \midrule
        \textbf{Average} 
        & \textbf{0.88} & \textbf{0.85} & \textbf{0.86} & \textbf{0.71} \\
        \bottomrule
    \end{tabular}
\end{table}

Table~\ref{tab:event_strategy_results} summarizes event localization performance on the manually annotated evaluation subset (162 segments, 94 minutes). Using a segment-level matching criterion with $\mathrm{tIoU}\ge 0.5$, the proposed pipeline achieves an average F1-score of 0.86 and a mean temporal IoU of 0.71 across all event categories.

Depth deviation and lens contamination events exhibit the strongest performance, with F1-scores of 0.89 and 0.92, respectively. These events are characterized by distinctive geometric or photometric signatures, making them easier to localize reliably. For depth deviation events, the estimated working-distance change achieves a mean absolute error (MAE) of 3.1~mm, which lies within the clinically relevant operating tolerance for laparoscopic camera adjustment.

Instrument--tissue interaction and visibility degradation events show comparatively lower recall and temporal overlap. These phenomena often emerge gradually and lack sharply defined temporal boundaries, leading to slight mismatches in onset and offset estimation. Nevertheless, the achieved F1-scores above 0.80 indicate that the event taxonomy and multi-cue detection strategy are sufficiently robust to capture camera-relevant surgical states.

The semantic consistency of the discovered strategy clusters is evaluated by comparing cluster assignments with surgeon-provided strategy labels on the same held-out subset. The clustering achieves a purity of 0.81 and a normalized mutual information (NMI) of 0.77, indicating strong agreement between automatically mined strategy structures and expert interpretations.

In addition, the mean intra-cluster variance of strategy-relevant attributes (depth deviation, local entropy change, and optical-flow directionality) is 0.46 averaged across all clusters. The low variance confirms that events grouped within the same cluster exhibit coherent geometric and visual patterns rather than arbitrary co-occurrence.

The quantitative results of this ablation are summarized in Fig.~\ref{fig:k_ablation}. 
Increasing $K$ from 8 to 12 improves semantic consistency (higher NMI) and reduces 
intra-cluster variance, indicating more compact and interpretable strategy grouping. 
For $K \ge 14$, the gains saturate and a slight increase in variance is observed, 
suggesting fragmentation into overly fine-grained clusters. Based on this trade-off 
between semantic coherence and structural compactness, we use $K=12$ as the default 
setting in the proposed framework.

These results support the hypothesis that expert camera control can be characterized by a limited set of reusable strategy primitives. By jointly considering temporal structure and attribute similarity, the WSBGC framework effectively suppresses demonstration-specific noise and extracts stable camera-handling regimes. The compact intra-cluster structure further ensures that the mined strategies provide consistent and interpretable supervision signals for downstream multi-modal policy learning.

\subsection{Ex Vivo Laparoscopic Camera-Control Evaluation}

\begin{figure*}[t!]
    \centering
    \includegraphics[width=1.0\textwidth]{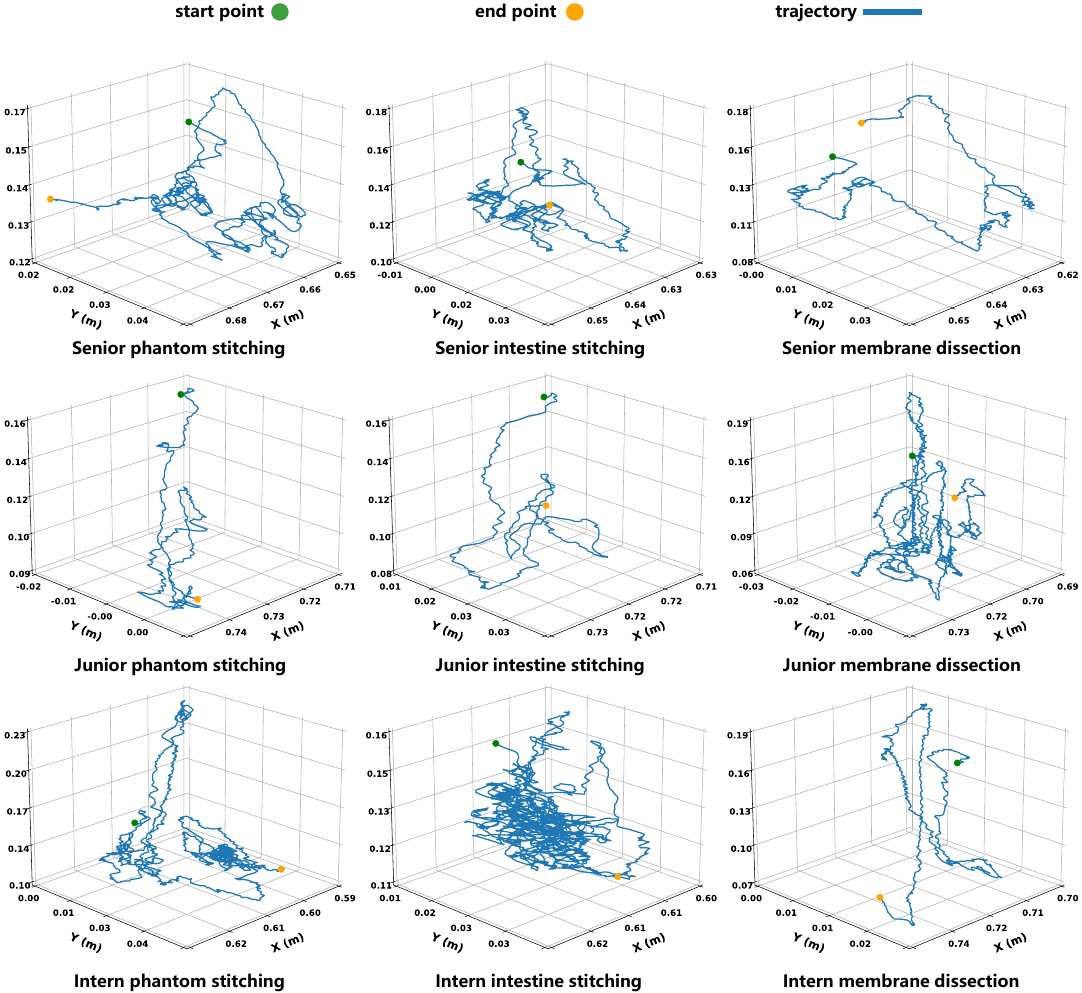}
    \caption{\textbf{3D spatial trajectories of the autonomous robotic laparoscopic system across diverse surgical tasks and operator expertise levels:} The trajectories illustrate the scope’s movement and the control system's performance during three benchmarking tasks: silicone stitching, intestine stitching, and dissection. Rows represent the performance of senior, junior, and intern surgeons, respectively. In each subplot, the path initiates from the start point (green dot) and concludes at the end point (orange dot). All spatial coordinates are measured in meters (m)}
    \label{fig:Experimental execution diagram}
\end{figure*}

The ex vivo surgical experiments are conducted to evaluate the performance of the proposed laparoscopic camera control method. All processed data is publicly on Zenodo at https://doi.org/10.5281/zenodo.18650476. As shown in Fig.~\ref{surgical-validation}, the functionalities of the proposed system are illustrated through porcine stomach membrane dissection and silicone phantom stitching tasks. In the multi-site dissection performed on porcine stomach tissue, the images demonstrate the autonomous following behavior of the laparoscope and its ability to dynamically adjust the penetration ratio in response to shifts in the surgical instrument workspace. This capability enables the system to maintain an appropriate visual field, thereby facilitating efficient task execution. In the simulated silicone stitching task, the system further demonstrates multi-modal interaction capability, allowing the surgeon to refine the laparoscope pose via verbal commands without interrupting the manual stitching process. 

As shown in Fig.~\ref{fig:autonomous-laparoscopic-clean-validation}, under mild fogging conditions, the laparoscope remains stationary, as such transient fogging is likely to dissipate naturally and is generally tolerable without compromising the surgical procedure. This design avoids unnecessary scope withdrawal that could disrupt the surgeon’s operation. In contrast, when severe fogging or lens contamination is detected, the system automatically initiates an “extraction–watering–wiping” sequence to restore visual clarity. The results demonstrate the robustness of the proposed framework in maintaining an optimal surgical field of view (FoV) through autonomous instrument withdrawal and lens cleaning.

\begin{figure*}[htbp]
    \centering
    \includegraphics[width=\linewidth]{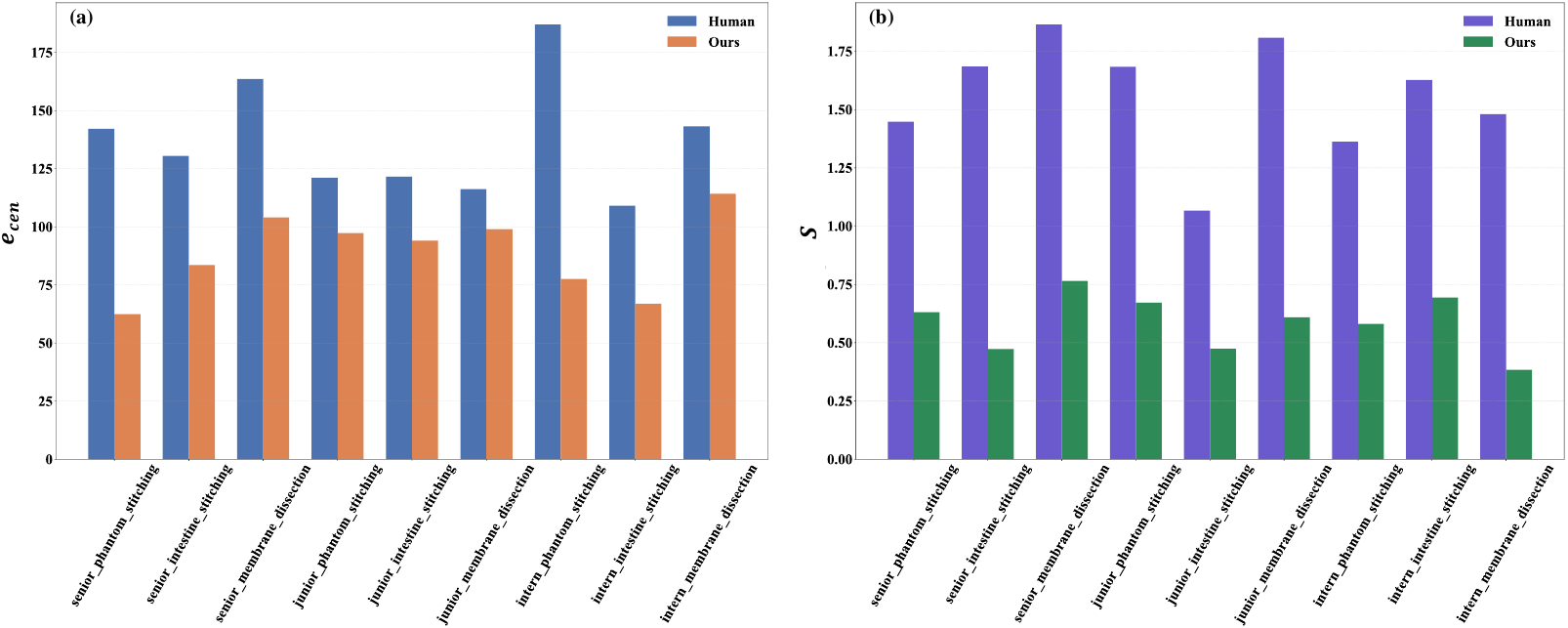}
    \caption{\textbf{Comparison of Field-of-View Centering and Image Shaking performance between manual camera control and the proposed method:} (a): The average pixel errors between the current and the desired instrument tip pixel coordinates under two methods. (b): The average pixel differences of the selected feature point between two adjacent images.}
    \label{Field-of-View-Centering-and-Image-shaking}
\end{figure*}

To quantify the effectiveness of the proposed method in stabilizing the laparoscopic view and providing a clear surgical scene for instrument manipulation and task execution, the Field-of-View Centering and Image Shaking metrics are first analyzed. For both manual camera control and the proposed method, the results of $e_{cen}$ and $S$ across the whole $9$ trials described in Section \ref{experimental-protocol} are summarized in Table. \ref{pixel-error-and-difference}. For a clear visualization of the performance differences, a graphical comparison is presented in Fig. \ref{Field-of-View-Centering-and-Image-shaking}. 

As shown in Table. \ref{pixel-error-and-difference}, the proposed autonomous controller reduced the overall centering error $e_{cen}$ by $35.26\%$ compared with manual camera control, indicating improved alignment of the surgical view. Moreover, the stability metric $S$ decreased by $62.33\%$, demonstrating a substantial enhancement in field-of-view (FoV) stability.

\begin{table*}[t]
    \centering
    \caption{Comparison of Field-of-View Centering and Image Shaking performance between manual camera control and the proposed method.}
    \label{pixel-error-and-difference}
    \normalsize

    \newcolumntype{Y}{>{\centering\arraybackslash}X}
    \setlength{\tabcolsep}{2.5pt}
    \renewcommand{\arraystretch}{1.15}
    \small
    
    \begin{tabularx}{\textwidth}{@{}
        >{\centering\arraybackslash}p{4.0cm}
        >{\centering\arraybackslash}p{2.5cm}
        *{9}{Y}
    @{}}
    \toprule
    \textbf{Metric} & \textbf{Method}
    & $1^{\mathrm{st}}$ & $2^{\mathrm{nd}}$ & $3^{\mathrm{rd}}$ & $4^{\mathrm{th}}$ & $5^{\mathrm{th}}$
    & $6^{\mathrm{th}}$ & $7^{\mathrm{th}}$ & $8^{\mathrm{th}}$ & $9^{\mathrm{th}}$ \\
    \midrule

    \multirow{2}{*}{\textbf{Field-of-View Centering: $e_{cen}$}}
    & \textbf{Human}
    & 142.12 & 130.45 & 163.61 & 121.02 & 121.54 & 116.17 & 187.00 & 109.09 & 143.25 \\
    & \textbf{Ours}
    & 62.47 & 83.52 & 104.01 & 97.35 & 94.07 & 99.04 & 77.59 & 66.86 & 114.20 \\
    \midrule

    \multirow{2}{*}{\textbf{Image Shaking: $S$}}
    & \textbf{Human}
    & 1.45 & 1.68 & 1.86 & 1.68 & 1.07 & 1.81 & 1.36 & 1.63 & 1.48 \\
    & \textbf{Ours}
    & 0.63 & 0.47 & 0.77 & 0.67 & 0.47 & 0.61 & 0.58 & 0.69 & 0.38 \\
    
    \bottomrule
    \end{tabularx}

    \vspace{2mm}
    \begin{minipage}{\textwidth}
        \footnotesize
        \textit{\textbf{Note:}} The units of the values are all pixels.
    \end{minipage}
\end{table*}

For the elapsed time interval $\delta t_{target}$ required for the instrument tip to reach the desired pixel coordinate from its current position, the proposed method converged in approximately $7.70 \, \mathrm{s}$. In contrast, manual camera control by a human assistant resulted in a shorter time-to-target. This behavior is expected, since the underlying IBVS controller requires a finite amount of time to converge to the desired image feature position
while human assistants can anticipate abrupt instrument motions and adjust the viewpoint more aggressively, resulting in faster transient responses. Nevertheless, the proposed autonomous method achieves smoother and more stable camera control, which is beneficial for maintaining a consistent surgical view during delicate tissue manipulation. 

Additionally, the accuracy of verbal modifier interpretation approaches $100\%$. During the trials, surgeons occasionally needed to repeat an instruction two or three times before it was correctly recognized. Nevertheless, no misinterpretation was observed across the $10$ trials. Once successfully recognized, the VLM consistently generated the correct command without producing erroneous outputs.

In terms of working stability, according to the positional trajectories shown in Fig. \ref{fig:Experimental execution diagram}, the overall high-frequency spectral energy ratio $\vec{R}$ is analyzed, and the average result of which over the $9$ trials is $\left[1.26 \times 10^{-4}, 8.06 \times 10^{-5}, 1.58 \times 10^{-4}\right]$. These extremely low values indicate that the majority of the spectral energy is concentrated in the low-frequency range. This suggests that the positional trajectory is predominantly smooth, with minimal high-frequency oscillatory components, thereby reflecting stable operational performance. On the other hand, the proposed autonomous controller achieved a mean absolute error $e_{depth}$ of $0.0437$ over the $10$ trials, with a mean relative error $\hat{e}_{depth}$ of $7.12\%$ in the context of working distance. The mean relative error is relatively small, i.e., $\leq 10\%$, indicating that the proposed method achieves accurate regulation of the laparoscope penetration ratio. This enables flexible adjustment of the working distance, thereby allowing the surgical view to be adapted more effectively to task requirements. On the other hand, the mean pose jitter $J$ is $4.33 \times 10^{-4} \, \mathrm{m/s}$, indicating highly smooth and stable camera motion throughout the surgical procedures. 


The average recovery time $\delta t_{recover}$ was $25 \,\mathrm{s}$, within which $10 \,\mathrm{s}$ were allocated for the actual cleaning procedure performed by the assistant or nurse. This duration provides sufficient time for effective laparoscope cleaning while maintaining an acceptable interruption to the surgical workflow.

\section{Conclusion}
In this work, we have established a strategy-supervised framework for autonomous laparoscopic camera control, bridging the gap between semantic surgical understanding and precise kinematic execution. By eschewing direct end-to-end action regression, our pipeline first parses surgical videos into interpretable event graphs to mine latent camera-handling primitives. These discovered strategies supervise a Vision-Language Model to predict discrete motion intentions, which are robustly executed by an IBVS-RCM controller. Our results confirm that this hierarchical approach yields highly stable closed-loop behavior. The system effectively handles diverse intraoperative events---such as depth changes and lens contamination---and achieves significant improvements in view stability (62.33\% reduction in shaking) and centering accuracy (35.26\% error reduction) compared to manual operation. Furthermore, the alignment between mined clusters and expert annotations (NMI 0.77) validates that our graph-based mining successfully captures the tacit knowledge of expert camera operators. While current validation relies on \textit{ex vivo} models and cholecystectomy data, future work will focus on expanding the strategy knowledge base across diverse surgical procedures and conducting \textit{in vivo} trials. This will allow us to further evaluate the system's robustness under physiological motion and its seamless integration into clinical workflows.

\section*{Acknowledgments}

{\appendices

\section{Camera-Response Motion Estimation}
\label{sec:camera_action}

This appendix details the computation of interval-level camera-response
motion descriptors used for strategy mining and directional supervision.
The objective is not to recover metric camera pose, but to obtain robust
viewpoint-relative motion descriptors that characterize compensatory
camera adjustments within each event interval.
For each consecutive frame pair $(I_t, I_{t+1})$, we estimate background
motion induced by camera movement while suppressing tool motion and
unreliable regions. The resulting frame-wise motion increments are
aggregated over event intervals to form 3-DoF motion descriptors
in the image space.

\paragraph{Background Mask Construction}

Let $M_{\mathrm{tool}}(t)$ denote the tool segmentation mask and
$M_{\mathrm{border}}$ the image border mask. Let
$M_{\mathrm{vis}}(t)$ denote a visibility-quality mask derived from
focus score $F(t)$ and contamination score $S_{\mathrm{cont}}(t)$.

The valid background region is defined as:
\begin{equation}
R_{\mathrm{bg}}(t)
=
\neg M_{\mathrm{tool}}(t)
\cap
\neg M_{\mathrm{border}}
\cap
M_{\mathrm{vis}}(t).
\end{equation}

Frames with insufficient valid pixels (e.g., heavy smoke or contamination)
are excluded from motion estimation.

\paragraph{Frame-Wise Dense Motion Estimation}

For valid frame pairs $(I_t, I_{t+1})$, we compute dense optical flow
$\mathbf{f}_t(u,v) = (f_u, f_v)$ using a pre-trained estimator.
The flow field is restricted to the background region:
\begin{equation}
\mathcal{F}_t
=
\left\{
\mathbf{f}_t(u,v)
\mid
(u,v) \in R_{\mathrm{bg}}(t)
\right\}.
\end{equation}

This suppresses motion induced by surgical tools and non-rigid tissue
interaction.

\paragraph{Robust Image-Plane Motion Fitting}

To extract camera-induced image-plane motion, we fit a 2D translation
model to $\mathcal{F}_t$ using RANSAC:
\begin{equation}
\mathbf{p}_{t+1}
\approx
\mathbf{p}_t
+
\begin{bmatrix}
\delta u_t \\
\delta v_t
\end{bmatrix},
\end{equation}
where $\delta u_t, \delta v_t$ represent image-plane translations.
RANSAC suppresses outliers caused by residual deformation
or local artifacts.

\paragraph{Axial Depth Motion}

Axial camera motion is derived independently using the filtered
working-distance trajectory $\tilde{z}(t)$ defined in
Section~\ref{sec:offline_event_parsing}:
\begin{equation}
\delta z_t = \tilde{z}(t+1) - \tilde{z}(t).
\end{equation}

This separates depth-induced scaling effects from in-plane motion.

\paragraph{Interval-Level Aggregation}

For an event interval $e_k = [t_s, t_e]$, cumulative motion
components are computed as:
\begin{equation}
\Delta u_k = \sum_{t=t_s}^{t_e-1} \delta u_t,
\quad
\Delta v_k = \sum_{t=t_s}^{t_e-1} \delta v_t,
\end{equation}
\begin{equation}
\Delta z_k = \tilde{z}(t_e) - \tilde{z}(t_s).
\end{equation}

The final interval-level camera-response descriptor is defined as:
\begin{equation}
\mathbf{x}^{\mathrm{action}}_k
=
(
\Delta u_k,\,
\Delta v_k,\,
\Delta z_k
).
\end{equation}

\section{Real-Time Laparoscopic Control (IBVS with RCM Constraint)}

\label{app:ibvs_rcm_control}

In MIS, the laparoscope is inserted through a trocar and must satisfy a strict
Remote Center of Motion (RCM) constraint: it is allowed to translate along its
longitudinal axis and rotate about the incision point. Meanwhile, the camera
must maintain a clear and stable view by keeping the surgical instrument tip
near the center of the field of view (FoV). We adopt the standard IBVS--RCM
control paradigm in \cite{2013-ICRA-RCM,2006-RAM-IBVS} and use it as the
low-level execution layer for the high-level directional commands predicted by
our vision model.

\paragraph{RCM constraint}
At time $t$, one end of the laparoscope is rigidly attached to the manipulator
end-effector (EE). The endpoint attached to the EE is denoted as
$\vec{p}_{i}\in\mathbb{R}^{3}$ and the opposite endpoint as
$\vec{p}_{i+1}\in\mathbb{R}^{3}$. The trocar point is
$\vec{p}_{trocar}\in\mathbb{R}^{3}$. The ratio between the external portion
length and the whole length of the laparoscope is $\lambda\in\mathbb{R}$.
The RCM point along the laparoscope shaft is
\begin{equation}
\label{RCM-constrain-formulation-1}
\vec{p}_{rcm}\!\left(\vec{q}(t)\right) =
\vec{p}_{i}\!\left(\vec{q}(t)\right) + \lambda(t)\!\left(\vec{p}_{i+1}\!\left(\vec{q}(t)\right)
- \vec{p}_{i}\!\left(\vec{q}(t)\right)\right),
\end{equation}
where $\vec{q}(t)\in\mathbb{R}^{n}$ denotes the joint configuration and $n$
the number of joints. Differentiating yields:
\begin{equation}
\label{RCM-constrain-formulation-2}
\dot{\vec{p}}_{rcm} = \dot{\vec{p}}_{i} + \dot{\lambda}\left(\vec{p}_{i+1}-\vec{p}_{i}\right)
+ \lambda\left(\dot{\vec{p}}_{i+1}-\dot{\vec{p}}_{i}\right),
\end{equation}
and using Jacobians $\mat{J}_{i},\mat{J}_{i+1}\in\mathbb{R}^{3\times n}$:
\begin{equation}
\label{RCM-constrain-formulation-3}
\dot{\vec{p}}_{rcm}
= \mat{J}_{i}\dot{\vec{q}} + \dot{\lambda}\left(\vec{p}_{i+1}-\vec{p}_{i}\right)
+ \lambda\left(\mat{J}_{i+1}\dot{\vec{q}} - \mat{J}_{i}\dot{\vec{q}}\right).
\end{equation}
Reorganizing gives:
\begin{equation}
\dot{\vec{p}}_{rcm} =
\begin{bmatrix}
\mat{J}_{i} + \lambda\left(\mat{J}_{i+1} - \mat{J}_{i}\right) \\[6pt]
\vec{p}_{i+1}-\vec{p}_{i}
\end{bmatrix}^\top
\begin{bmatrix}
\dot{\vec{q}} \\[6pt]
\dot{\lambda}
\end{bmatrix}
=
\mat{J}_{rcm}\left(\vec{q}, \lambda\right)
\begin{bmatrix}
\dot{\vec{q}} \\[6pt]
\dot{\lambda}
\end{bmatrix}.
\end{equation}
The RCM constraint enforces $\vec{p}_{rcm}\equiv \vec{p}_{trocar}$, thus
$\dot{\vec{p}}_{rcm}=\vec{0}$.

\paragraph{IBVS objective}
Let the pixel coordinate of the instrument tip be $\vec{f}\in\mathbb{R}^{2}$ and
its depth be $z\in\mathbb{R}$. The feature velocity is related to the camera
twist $\vec{\xi}_c=[\vec{v}_c,\vec{\omega}_c]^\top\in\mathbb{R}^{6}$ through the
interaction matrix $\mat{J}_{int}(\cdot)\in\mathbb{R}^{2\times6}$:
\begin{equation}
\label{IBVS-formulation-1}
\dot{\vec{f}} = \mat{J}_{int}\left(\vec{f}_{d}, z_{d}\right)
\begin{bmatrix}
\vec{v}_c \\[6pt]
\vec{\omega}_c
\end{bmatrix}.
\end{equation}
The interaction matrix at the desired configuration is:
\begin{align}
\label{IBVS-formulation-2}
&\mat{J}_{int}\left(\vec{f}_{d}, z_{d}\right) \\[6pt] \notag
&=
\begin{bmatrix}
-\dfrac{1}{z_{d}} & 0 & \dfrac{x_{d}}{z_{d}} & x_{d}y_{d} & -(1+{x_{d}}^2) & y_{d} \\[6pt]
0 & -\dfrac{1}{z_{d}} & \dfrac{y_{d}}{z_{d}} & 1+{y_{d}}^2 & -x_{d}y_{d} & -x_{d}
\end{bmatrix},
\end{align}
with normalized desired coordinates
\begin{equation}
\label{IBVS-formulation-3}
\vec{\mathcal{F}}^{*}_{I} = \mat{K}^{-1} \vec{f}_{d},
\end{equation}
where $\mat{K}\in\mathbb{R}^{3\times3}$ is the camera intrinsic matrix.

The camera twist is related to joint velocities via the camera Jacobian
$\mat{J}_c(\vec{q})\in\mathbb{R}^{6\times n}$:
\begin{equation}
\label{IBVS-RCM-formulation-1}
\dot{\vec{f}} = \mat{J}_{int}\left(\vec{f}_{d}, z_{d}\right)\mat{J}_c\left(\vec{q}\right)\dot{\vec{q}}
= \mat{J}_{v}\left(\vec{f}_{d}, z_{d}, \vec{q}\right)\dot{\vec{q}},
\end{equation}
where
\begin{equation}
\label{IBVS-RCM-formulation-2}
\mat{J}_c = \begin{bmatrix}
\mat{R}^\top_{c,ee} & -\mat{R}^\top_{c,ee} \left[\vec{p}_{c,ee}\right]_\times \\[6pt]
\mat{0}_{3 \times 3} & \mat{R}^\top_{c,ee}
\end{bmatrix}
\begin{bmatrix}
\mat{R}^\top_{ee,w} & \mat{0}_{3 \times 3} \\[6pt]
\mat{0}_{3 \times 3} & \mat{R}^\top_{ee,w}
\end{bmatrix}
\mat{J}_{r}.
\end{equation}

\paragraph{Augmented IBVS--RCM task}
Combining the IBVS objective with the RCM constraint yields the augmented task
derivative $\dot{\vec{t}}_{aug}\in\mathbb{R}^{5}$ and augmented Jacobian
$\mat{J}\in\mathbb{R}^{5\times(n+1)}$:
\begin{align}
\label{IBVS-RCM-formulation-3}
\dot{\vec{t}}_{aug}
=
\begin{bmatrix}
\dot{\vec{f}} \\[6pt]
\mat{0}_{3 \times 1}
\end{bmatrix}
=
\mat{J}
\begin{bmatrix}
\dot{\vec{q}} \\[6pt]
\dot{\lambda}
\end{bmatrix}
=
\begin{bmatrix}
\mat{J}_v & \mat{0}_{2 \times 1} \\[6pt]
\multicolumn{2}{c}{\mat{J}_{rcm}}
\end{bmatrix}
\begin{bmatrix}
\dot{\vec{q}} \\[6pt]
\dot{\lambda}
\end{bmatrix}.
\end{align}
The desired augmented task is $\vec{t}_{d} = [\vec{f}_{d}, \vec{p}_{trocar}]^\top$
and the task error is:
\begin{equation}
\label{IBVS-RCM-formulation-4}
\vec{e}_{t} =
\begin{bmatrix}
\vec{f}_{d} - \vec{f} \\
\vec{p}_{trocar} - \vec{p}_{rcm}
\end{bmatrix}.
\end{equation}
The feedback control law is:
\begin{equation}
\label{IBVS-RCM-formulation-5}
\begin{bmatrix}
\dot{\vec{q}} \\[6pt]
\dot{\lambda}
\end{bmatrix}
=
\mat{J}^{\#}
\begin{bmatrix}
\mat{K}_f & \mathbf{0}_{2 \times 3} \\[6pt]
\mathbf{0}_{3 \times 2} & \mat{K}_{rcm}
\end{bmatrix}
\vec{e}_t + \left(\mat{I} - \mat{J}^{\#} \mat{J}\right)\vec{\omega},
\end{equation}
where $\mat{K}_{f}\in\mathbb{R}^{2\times2}$ and
$\mat{K}_{rcm}\in\mathbb{R}^{3\times3}$ are positive definite diagonal gain
matrices, $\mat{J}^{\#}$ is the pseudoinverse, and $\vec{\omega}$ is a null-space
input.

\paragraph{Directional command execution from the vision model}
Our vision model predicts a discrete 3-DoF directional command
$\hat{\mathbf{d}}_t\in\{-1,0,+1\}^{3}$,
defined in the image-based command space as
$\hat{\mathbf{d}}_t = (\hat{d}_u,\hat{d}_v,\hat{d}_z)$.
This command is converted into incremental setpoints for the IBVS--RCM
controller, which then generates feasible end-effector pose updates under the
trocar pivot constraint in closed loop.

Specifically, lateral directions are mapped to small pixel-space shifts of the
desired feature position:
\begin{equation}
\vec{f}_{d}(t) = \vec{f}(t) + \Delta \vec{f}\bigl(\hat{d}_u,\hat{d}_v\bigr),
\end{equation}
where
\begin{equation}
\Delta \vec{f}\bigl(\hat{d}_u,\hat{d}_v\bigr)=
\begin{bmatrix}
s_u \hat{d}_u\\
s_v \hat{d}_v
\end{bmatrix},
\end{equation}
and $\hat{d}_u,\hat{d}_v\in\{-1,0,+1\}$ encode left/right and up/down
corrections, respectively. Longitudinal commands are mapped to penetration
updates:
\begin{equation}
\lambda_{d}(t) = \lambda(t) + \Delta \lambda\bigl(\hat{d}_z\bigr),
\qquad
\Delta \lambda\bigl(\hat{d}_z\bigr)= s_z \hat{d}_z,
\end{equation}
with $s_u,s_v$ (pixels) and $s_z$ (mm) fixed step sizes. The discrete command
selects the sign of the setpoint shift, while the resulting motion magnitude is
governed by the IBVS feedback law under the RCM constraint. In all cases,
commands are executed at a fixed rate and are safety-filtered by joint,
velocity, and workspace limits.

\paragraph{Penetration regulation}
To regulate the penetration degree, we define the cost:
\begin{equation}
\label{IBVS-RCM-formulation-6}
d = \frac{1}{2} \| \lambda - \lambda_{d}\|^2,
\end{equation}
and choose the last element of $\vec{\omega}$ as the negative gradient to drive
$\lambda$ toward $\lambda_{d}$:
\begin{equation}
\vec{\omega} = -\left(\nabla_{\left(\vec{q},\lambda\right)} d\right)^\top
= \left(0, \cdots\ 0, \lambda_{d} - \lambda\right)^\top.
\end{equation}
With Eq.~(\ref{IBVS-RCM-formulation-5}), the laparoscope pose is controlled by
the desired pixel coordinate $\vec{f}_{d}$ and the penetration setpoint
$\lambda_{d}$, while satisfying the RCM constraint.

}




 





\bibliography{ref.bib}
\end{document}